%% file: acl2024_conference.tex
\pdfoutput=1

\documentclass[11pt]{article}

\usepackage{acl}

\usepackage{times}
\usepackage{latexsym}

\usepackage[T1]{fontenc}

\usepackage[utf8]{inputenc}

\usepackage{microtype}

\usepackage{inconsolata}

\usepackage{hyperref}
\usepackage{url}
\usepackage{amsmath}
\usepackage{graphicx}
\usepackage{color}
\usepackage{xcolor}
\usepackage{siunitx}
\usepackage{amsfonts}
\usepackage{latexsym}
\usepackage{comment}
\usepackage{booktabs}
\usepackage{microtype}
\usepackage{xspace}
\usepackage{booktabs}
\usepackage{multirow}
\usepackage{dcolumn}
\usepackage{tikz}
\usepackage{bbding}
\usepackage{hhline}
\usepackage{mdframed}
\usepackage{enumitem}
\usepackage{hyperref}
\usepackage{url}
\newcolumntype{d}[1]{D{.}{.}{#1}}
\usepackage{makecell}
\usepackage{dirtytalk}
\usepackage{colortbl}
\usepackage{wrapfig}
\usepackage{caption}
\usepackage{bm}
\usepackage{subfigure}

\usepackage{soul}

\hypersetup{
  colorlinks   = true, 
  urlcolor     = blue!50!black, 
  linkcolor    = blue!50!black, 
  citecolor   = blue!50!black 
}

\title{What Evidence Do Language Models Find Convincing?}

\newcommand{\tightparagraph}[1]{
    \smallskip
    \noindent
    \textbf{#1}
}

\definecolor{green}{rgb}{0.1,0.1,0.1}
\definecolor{chocolate}{HTML}{D2691E}
\definecolor{maroon}{HTML}{A00000}
\definecolor{indigo}{HTML}{4B0082}
\definecolor{green}{HTML}{008000}
\definecolor{red}{HTML}{a91e1e}
\definecolor{cadmiumgreen}{rgb}{0.0, 0.42, 0.24}
\definecolor{forestgreen}{rgb}{0.13, 0.55, 0.13}

\DeclareMathOperator{\E}{\mathbb{E}}
 
\newcolumntype{L}[1]{>{\PreserveBackslash\raggedright}p{#1}}
\newcolumntype{R}[1]{>{\raggedleft\let\newline\\\arraybackslash\hspace{0pt}}m{#1}}
\newcolumntype{P}[1]{>{\centering\arraybackslash}p{#1}}

\usepackage{amssymb}
\usepackage{pifont}

\makeatletter
\newcommand*\myfontsize{%
  \@setfontsize\myfontsize{8}{9}%
}
\makeatother

\newcommand{\dataname}{\textsc{ConflictingQA}}

\usepackage{adjustbox}

\usepackage{dsfont}

\author{\makecell{Alexander Wan, Eric Wallace, Dan Klein} \\
UC Berkeley \\
\texttt{\{alexwan, ericwallace, klein\}@berkeley.edu} \\
}

\begin{document}
\maketitle

\begin{abstract}
Retrieval-augmented language models are being increasingly tasked with subjective, contentious, and conflicting queries such as ``is aspartame linked to cancer''. 
To resolve these ambiguous queries, one must search through a large range of websites and consider \emph{which, if any, of this evidence do I find convincing?}
In this work, we study how LLMs answer this question. 
In particular, we construct \dataname{}, a dataset that pairs controversial queries with a series of real-world evidence documents that contain different facts (e.g., quantitative results), argument styles (e.g., appeals to authority), and answers (\texttt{Yes} or \texttt{No}). We use this dataset to perform sensitivity and counterfactual analyses to explore which text features most affect LLM predictions. Overall, we find that current models rely heavily on the \textit{relevance} of a website to the query, while largely ignoring \textit{stylistic} features that humans find important such as whether a text contains scientific references or is written with a neutral tone.
Taken together, these results highlight the importance of RAG corpus quality (e.g., the need to filter misinformation), and possibly even a shift in how LLMs are trained to better align with human judgements.
\end{abstract}

\section{Introduction}\label{sec:intro}\input{sections/01-intro}
\section{Background and Motivations}\label{sec:background}\input{sections/02-background}
\section{The \dataname{} Dataset}\label{sec:data}\input{sections/03-data}
\section{Experimental Results}\label{sec:exp}\input{sections/04-exp}

\section{Discussion \& Related Work}\label{sec:discuss}\input{sections/05-discuss}
\section{Conclusion}\label{sec:concl}\input{sections/06-concl}

\subsubsection*{Acknowledgments} We thank Sewon Min and the members of Berkeley NLP for useful feedback on this project. We thank Ruiqi Zhong for helping us run the D5 codebase. Eric Wallace is supported by the Apple Scholars in AI/ML Fellowship.
Part of this research was supported with Cloud TPUs from Google’s TPU Research Cloud (TRC).

\bibliography{acl.bib}

\clearpage
\appendix
\input{sections/10-app}

\end{document}

%% file: sections/01-intro.tex
\begin{figure*}[ht]
\centering
\includegraphics[trim={0.1cm, 7.05cm, 0.1cm, 0.1cm}, clip, width=\textwidth]{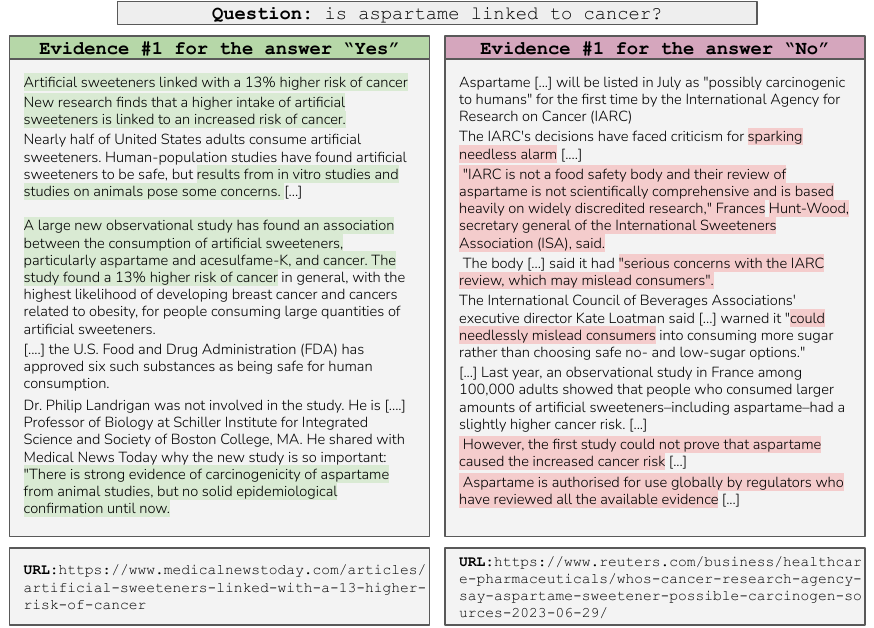}
\vspace{-0.51cm}
\caption{In \dataname{}, we create contentious questions such as ``\textit{is aspartame linked to cancer}''. We also retrieve evidence paragraphs for each question that contain different types of facts (e.g., quantitative results), argument styles (e.g., appeals to authority), and answers (\texttt{Yes} or \texttt{No}). For example, in the figure above we show two evidence paragraphs with their key arguments highlighted. Using \dataname{}, we study \textit{why} LLMs trust certain types of evidence paragraphs and argument styles over others.}
\label{fig:overview}
\end{figure*}

LLMs are widely deployed in settings that require understanding context---from retrieval-augmented systems to web agents, models condition on sources that range from internet paragraphs~\citep{karpukhin2020dense} to Python interpreters~\cite{gao2023pal}. At the same time, today's models are also given tasks that are increasingly open-ended and controversial, such as ``tell me if aspartame causes cancer''. To answer such questions, LLMs will read real-world paragraphs that are  contradictory, noisy, and rife with misinformation~\citep{bush2019bing}.

Humans have techniques to sift through such large quantities of complex and contradictory evidence by answering the question: \emph{which, if any, of this evidence do I find convincing?} To do so, humans combine multiple strategies, including fact checking and evaluating a source's credibility~\citep{fogg2003}, harnessing prior knowledge and beliefs~\citep{kakol2017c3}, and critically evaluating logical arguments~\citep{Metzger2010SocialAH}.

In this work, we explore how LLMs resolve similar ambiguities when faced with conflicting open-ended questions. To study this, we create \dataname{}, a dataset consisting of questions and real web documents that lead to conflicting answers. For instance, in Figure~\ref{fig:overview} we show an example where we pair the question ``is aspartame linked to cancer'' with a series of conflicting evidence documents collected from Google search. In our experiments, we evaluate the \textit{convincingness} of each evidence document by computing the rate at which a model's predictions align with that document's viewpoint (i.e., its win-rate).

We perform sensitivity and counterfactual analyses to find in-the-wild features that correlate with document convincingness. We consider a mix of features that describe (1) stylistic properties of a document and (2) the relevance of a document to the question. Many of these were inspired by results from studies on human credibility judgements: for example, we consider whether adding scientific references makes text appear more convincing.

Overall, we find that stylistic features play a considerably less impactful role in determining the convincingness of text than measures of relevance. Notably, we show that a simple perturbation targeting a website's relevance---prefixing the page with ``The following text is about the question: \texttt{[question]}''---is enough to substantially improves its win-rate. On the other hand, stylistic features like the informational content, whether a page contains references, or its confidence, tend to only have a neutral to negative effect on win-rate. These results show that LLM perceptions of convincingness, when grounded in real-world QA tasks, do not align with humans. Taken together, these results suggest there should be an increasing focus on the \textit{quality} of retrieved evidence and a shift in how LLMs are trained to align with human preferences. We release our code at \url{https://github.com/AlexWan0/rag-convincingness}.

%% file: sections/02-background.tex
\begin{figure*}[t]
\begin{center}
    \includegraphics[trim={0.1cm, 0.1cm, 0.1cm, 0.1cm}, clip, width=0.80\textwidth]{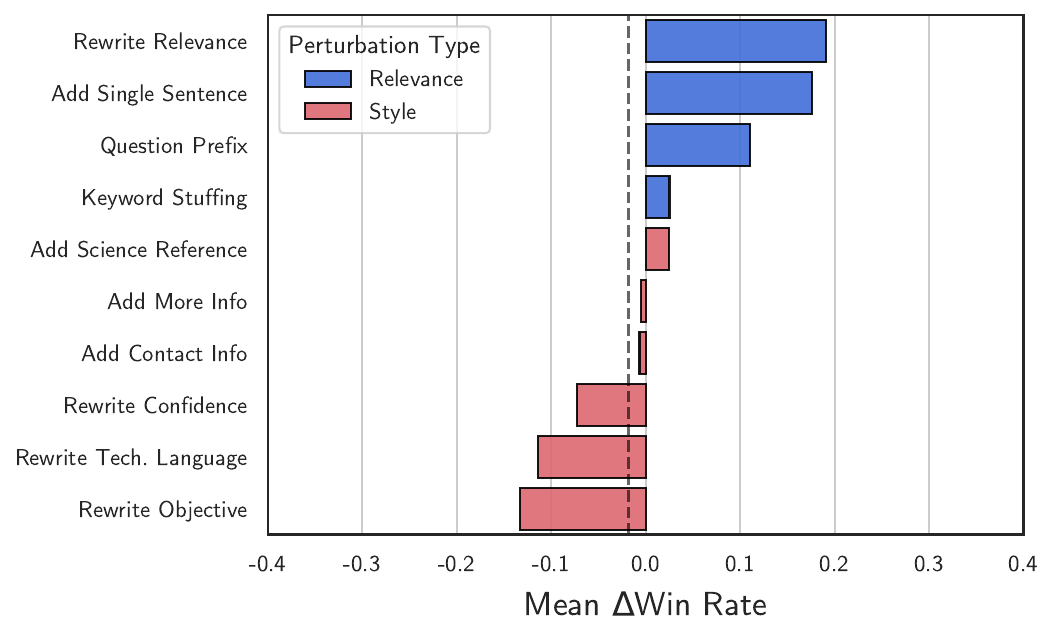}
\end{center}
\vspace{-0.48cm}
\caption{\textit{Models over-rely on document relevance}. We study how the convincingness of a particular evidence paragraph (measured through win-rate) changes when we modify it. We compare the effects of these changes to a baseline perturbation where we append ``Thanks for reading!'' to the end of the text (indicated by the dotted line). We find that many stylistic changes---inspired by factors that influence humans---have a neutral or even negative effect on models. On the other hand, perturbations that increase the text's relevance but minimally change its style have a substantial positive effect on models. Descriptions for each perturbation can be found in Appendix~\ref{app:perturb_descriptions}.}
\label{fig:counterfactual_barchart}
\end{figure*}

Standard LLMs can be used to solve tasks that do not require context, e.g., writing basic Python code or answering simple trivia questions~\citep{gpt3,raffel2020exploring,touvron2023llama}. To give these models more knowledge, agency, and capabilities, recent efforts have augmented LLMs with retrieval~\citep{guu2020retrieval,karpukhin2020dense}, domain-specific tools~\citep{schick2023toolformer,gao2023pal,mialon2023augmented}, or even generic web access~\citep{nakano2021webgpt,adept,AutoGPT}. These enhancements allow LLMs to answer more challenging open-domain questions (e.g., ``is aspartame linked to cancer?'') or accomplish open-ended tasks (e.g., ``buy me a size 9 pair of blue running shoes'').

\tightparagraph{Handling conflicting evidence.} A key question is how retrieval-augmented LLMs handle scenarios where their context is conflicting, ambiguous, or uncertain.
There has been a large body of work that studies how \textit{humans} handle such conflicting evidence using HCI studies~\cite{fogg2003,kakol2013subjectivity,flanagin2000credibility,Metzger2010SocialAH,kakol2017c3} or by trying to predict human argument preferences~\cite{gleize2019you, toledo2019automatic, gretz2019largescale}, but little work has been done on evaluating how AI models handle such conflicts.

The existing work in AI has focused on conflicts between facts learned during pre-training and the evidence given during inference, finding that models are largely receptive to retrieved samples~\citep{longpre2021entity,xie2023adaptive,chen2022rich}. However,  these works focus on restricted settings such as QA over Wikipedia, where there are relatively uncontroversial factoid questions that have trusted evidence paragraphs.
Moreover, they do not focus on \textit{what types} of evidence models prefer. Our goal is to design a more realistic question answering benchmark to better analyze features about the evidence itself.

%% file: sections/03-data.tex
\makeatletter
 \def\SOUL@hlpreamble{%
 \setul{}{2.4ex}%
 \let\SOUL@stcolor\SOUL@hlcolor
 \SOUL@stpreamble
 }
\makeatother

\newcommand{\hlc}[2][yellow]{{%
    \colorlet{foo}{#1}%
    \sethlcolor{foo}\hl{#2}}%
}

\definecolor{A}{HTML}{FFC0C0}      
\definecolor{B}{HTML}{FFDAC0}        
\definecolor{C}{HTML}{FFFDC0}      
\definecolor{D}{HTML}{D0FFC0}          
\definecolor{E}{HTML}{C0FFFD}     
\definecolor{F}{HTML}{C0D0FF}            
\definecolor{G}{HTML}{C0C0FF}             
\definecolor{H}{HTML}{DAC0FF}          
\definecolor{I}{HTML}{FFC0FF}       
\definecolor{J}{HTML}{FFC0DA}      

\newcommand{\Biodiversity}{\hlc[C]{Biodiversity}}
\newcommand{\OnlineLearning}{\hlc[B]{Online Learning}}
\newcommand{\WebDesign}{\hlc[D]{Web Design}}
\newcommand{\Pharmacology}{\hlc[A]{Pharmacology}}
\newcommand{\Sustainability}{\hlc[E]{Sustainability}}
\newcommand{\Philosophy}{\hlc[F]{Philosophy}}
\newcommand{\NuclearEnergy}{\hlc[G]{Nuclear Energy}}
\newcommand{\WorkLifeBalance}{\hlc[H]{Work-Life Balance}}
\newcommand{\Somnology}{\hlc[I]{Somnology}}
\newcommand{\Biomechanics}{\hlc[J]{Biomechanics}}

\begin{table*}[!t]
\centering
\footnotesize
\begin{tabular}{lp{8cm}c}
\toprule
{\bf \shortstack{Category}} & {\bf \multirowcell{1}{Example Question}} & {\shortstack{\textbf{Num Evidence Docs}}} \\
\midrule
\arrayrulecolor{black!30}
\Pharmacology{}	& Are antidepressants more effective than placebo? & 10 \\
\midrule
\OnlineLearning{} & Are online degrees valued less by employers? & 10 \\
\midrule
\Biodiversity{} & Are bees the most important pollinators? & 10 \\
\midrule
\WebDesign{}	& Does longer website content rank better on Google? & 13 \\
\midrule
\Sustainability{} & Are electric cars really green? & 9 \\
\midrule
\Philosophy{} & Are humans fundamentally good or evil? & 7 \\
\midrule
\NuclearEnergy{} & Can nuclear power solve climate change? & 7 \\
\midrule
\WorkLifeBalance{} & Is unlimited vacation time beneficial for employees? & 10 \\
\midrule
\Somnology{} & Do older people need less sleep? & 8 \\
\midrule
\Biomechanics{} & Do compression garments improve athletic performance? & 13 \\
\arrayrulecolor{black}
\bottomrule
\end{tabular}
\vspace{-0.1cm}
\caption{In \dataname{}, we create controversial questions for 191 different categories (see Table~\ref{tab:full_categories} in Appendix~\ref{app:details} for the complete list). Above, we show an example question for ten different categories, as well as the number of evidence paragraphs used for each question when evaluating LLaMA-2 Chat.}
\label{table:examples}
\end{table*}

Here, we describe the construction of \dataname{}, our dataset that evaluates what types of evidence are convincing for LLMs.
We design \dataname{} to emulate the common setup for deploying retrieval-augmented LLMs: we retrieve the most relevant documents for a particular user query and place them in the LLM's context window~\citep{chen2017reading,shi2023replug,ram2023context}.
To build our dataset, we tackle three challenges: collecting contentious questions, identifying relevant and diverse evidence paragraphs, and grouping evidence paragraphs together to create conflicting examples.

\tightparagraph{Collecting contentious questions.} We first create a series of realistic open-ended questions for which there exists conflicting evidence online. Critically, unlike past work on ambiguity in QA~\citep{min2020ambigqa,zhang2021situatedqa,sun2023answering}, we want to collect \textit{unambiguous} questions that still have answer conflicts. For example, in Figure~\ref{fig:overview}, we show a question ``are artificial sweeteners linked to cancer?'', which is a widely-debated query in which there exist websites that support both answers. We design the questions to elicit binary responses of \texttt{Yes} or \texttt{No} to simplify evaluation.

We create questions using GPT-4. To ensure that the model generates a diverse set of questions we take inspiration from previous work in synthetic dataset generation~\citep{gunasekar2023textbooks, eldan2023tinystories} and stratify the generations by topic: we first generate question categories (e.g., climate change, robotics, oncology) then generate sets of questions conditioned on each category (full prompt provided in Table~\ref{tab:gpt4_prompt} in Appendix~\ref{app:details}). We qualitatively find that the questions are diverse and challenging; we show ten examples of them in Table~\ref{table:examples}. We additionally manually remove duplicate questions in the dataset.

\tightparagraph{Collecting evidence paragraphs.} Given these questions, we want to find evidence paragraphs that support both the answers of \texttt{Yes} and \texttt{No}. We also want these paragraphs to (1) contain a diverse range of argument styles, factual information, etc., and (2) be realistic inputs to an LLM. To handle this, we emulate running an real-world retrieval-augmented LLM system that uses the Google Search API as its retrieval engine. Concretely, we take the user's query, reformulate it, and take the top-$k$ results from Google search for the answer \texttt{Yes} and the top-$k$ results for the answer \texttt{No}.

\begin{table*}[!t]
\centering
\footnotesize
\begin{minipage}[t]{0.53\textwidth}
\centering
\begin{tabular}{lr}
\toprule
 Number of questions & 238 \\
 Number of question categories & 144 \\
\midrule 
 Number of retrieved paragraphs & 2,208 \\
 Average paragraph length (words) & 365.01 \\
\midrule
 Number of paragraphs with $\geq 5$ comparisons & 912 \\
 Average number of comparisons per paragraph & 6.54 \\
\bottomrule
\end{tabular}
\vspace{-0.1cm}
\caption{Basic statistics for \dataname{} when evaluating LLaMA-2 Chat. We start by collecting a set of controversial questions for different categories (top). For each question, we retrieve a series of paragraphs from a variety of domains (middle). To determine the convincingness of a paragraph, we compare it against at least five different paragraphs that have the opposite stance/viewpoint (bottom).}
\label{table:data_stats}
\end{minipage}
\hfill 
\begin{minipage}[t]{0.44\textwidth}
\centering
\begin{tabular}{lr}
\toprule
{\bf \shortstack{Domain}} & \bf \shortstack{Count} \\
\midrule
.com & 527 \\
.org & 175 \\
.gov & 59 \\
.edu & 57 \\
.net & 12 \\
\midrule
\# unique & 39 \\
\bottomrule
\end{tabular}
\vspace{-0.1cm}
\caption{The top five most common top-level domains found in \dataname{} for evaluating LLaMA-2 Chat. The dataset consists of a diverse range of sources, including organizations (\texttt{.org}), schools (\texttt{.edu}), and governments (\texttt{.gov}).}
\label{table:tld_stats}
\end{minipage}
\end{table*}

We first turn each question into affirmative and negative statements using GPT-3.5 Turbo, e.g., the question ``is aspartame safe?'' is converted to ``Aspartame is safe. Aspartame isn't harmful.'' and ``Aspartame is harmful. Aspartame isn't safe.''. We also put double quotes (to indicate to Google Search that we have exact-match keywords) around any tokens that do not change after rephrasing the question into either statement (e.g., ``aspartame''). For both the affirmative and negative statements,\footnote{We find that there's significant overlap between the webpages returned by the affirmative and negative rephrasings of the query. In future iterations, it would be reasonable to perform searches for only the affirmative \textit{or} negative statements.} we search the queries using the Google Search API and retrieve the top-$k$ documents.\footnote{We set $k = 20$ because qualitatively the relevancy of the results dropped off significantly after this point.} As is common in many retrieval-augmented models~\citep{nakano2021webgpt}, we do not consider any visual features of the web page. Instead, we extract the raw text from each document using jusText.\footnote{Package available at \url{https://github.com/miso-belica/jusText}. Although humans use visual features when considering the credibility and trustworthiness of a source~\cite{kakol2017c3,fogg2003}, we do not consider these features as most state-of-the-art LLMs do not use visual inputs.} Additionally, we do not explicitly include metadata like source URL, publication date, or page headings.

When searching queries such as ``aspartame is safe'', we still retrieve documents that argue that aspartame is unsafe. To label the document's actual stance, we use an ensemble of \texttt{claude-instant-v1} and \texttt{GPT-4-1106-preview} and keep only the samples where the two models agree (see Table~\ref{tab:stance_prompt} in Appendix~\ref{app:details} for the prompts).\footnote{After identifying the stance, we also feed the paragraphs into the downstream LLM that we are testing and make sure that its answer aligns with the paragraphs' predicted stances. This further filters the data, accounting for mistakes in the downstream model. See Appendix~\ref{app:clean} for details.}
Furthermore, we allow the LLM to say that a document is irrelevant to the query; if so, we also filter it from the input.

Finally, we want to isolate \textit{paragraphs} from these larger documents to feed into the LLMs (as is common in RAG systems). To do this, we extract the most relevant 512 token window of text inside the document. We run the TAS-B model~\cite{Hofstaetter2021_tasb_dense_retrieval} across windows of 512 tokens with a 256 token stride, compute the dot product between the model's embedding of that window and the model's embedding of the question, and take the highest scoring window. We filter out any documents whose highest-scoring window has a dot product below 95.

\tightparagraph{Creating conflicting examples.} The end result of our data collection process is (1) a set of controversial questions that (2) have evidence paragraphs which contradict one another. This data can be used in a variety of ways to ``stress test'' RAG systems in order to understand how they behave under conflicting scenarios. One example of this is shown in Figure~\ref{fig:overview}, and the subsequent section will explore numerous possible uses of \dataname{}.
Table~\ref{table:data_stats} and Table~\ref{table:tld_stats} present basic statistics for our final data, accounting for specific filtering done for LLaMA-2 Chat.

%% file: sections/04-exp.tex
In this section, we use \dataname{} to evaluate what types of evidence models find convincing. 

\subsection{Convincingness as Paragraph Win Rate}
\label{sub:winrate}

We mainly focus on using \dataname{} in a setup where we ask an LLM a question while providing two conflicting evidence paragraphs (one that supports \texttt{Yes} and one that supports \texttt{No}). Then, we measure which paragraph the model's answer aligns with. By repeating this for all pairs of paragraphs, we can define the convincingness of a particular paragraph as its \textit{win-rate}, i.e., what percent of the time a model picks the answer in that paragraph over the other paragraphs.

Concretely, let $\mathcal{P}_{q,s}$ be the set of top-$k$ paragraphs corresponding to a controversial question $q$ with stance $s \in \{\text{\texttt{yes}}, \text{\texttt{no}}\}$. We take an LLM $f$ (e.g., LLaMA-2 Chat) and ask it for a binary prediction for the question $q$, based on two paragraphs selected from the larger set, $p_\text{\texttt{yes}} \in \mathcal{P}_{q,\text{\texttt{yes}}}$ and $p_{\text{\texttt{no}}} \in \mathcal{P}_{q,\text{\texttt{yes}}}$. The model makes a prediction: $f(p_\text{\texttt{yes}}, p_{\text{\texttt{no}}}, q) \in \{\text{\texttt{yes}}, \text{\texttt{no}}\}$.

For each paragraph, we define its win-rate as the empirical probability of the model's prediction aligning with its stance when paired with a set of conflicting paragraphs, i.e.,
\begin{align}
    \nonumber
    \text{WR}(p_{\text{\texttt{yes}}}, q) =  \E_{p \sim \mathcal{P}_{q, \text{\texttt{no}}}} [ \mathds{1} [ f(p_{\text{\texttt{yes}}}, p, q) = \text{\texttt{yes}} ] ]
\end{align}
Finally, as the ordering of the retrieved evidence is known to bias model predictions \citep{xie2023adaptive}, we calculate win-rate based on both orderings of the retrieved paragraphs. We additionally filter our dataset to ensure that each win-rate calculation consists of comparisons with at least five unique paragraphs.

\tightparagraph{Models Cannot Predict Convincingness.} We designed the above experimental setting to emulate how production RAG models work. However, we could have instead just directly asked the LLM, ``\textit{do you find paragraph X to be persuasive?}''. This is how humans are typically asked to judge the convincingness of a piece of evidence~\citep{kakol2017c3, Jo2019HowDH, kakol2013subjectivity}. However, we find that LLMs are largely incapable of expressing the convincingness of a paragraph in words, e.g., there is little correlation in which paragraphs are marked as convincing in the two settings (Figure~\ref{fig:ungrounded_vs_winrate}).\footnote{Our methodology for this setting is described in more detail in Appendix~\ref{sec:ungrounded_method}.} We thus focus on the more practically-grounded setting going forward.

\subsection{Implementation Details}

We evaluate a mix of open-source (LLaMA-2 Chat \citep{touvron2023llama2}, Vicuna v1.5 \citep{vicuna}, and WizardLM v1.2 \citep{xu2023wizardlm}) and closed source (GPT-4, Anthropic Claude v1 Instant) models, extracting binary \texttt{Yes}/\texttt{No} predictions for each question. Importantly, we specify ``Use only the information in the above text to answer the question'' as we are looking to see how models judge stylistic differences in evidence, rather than for their prior stances on the question. For open-source models, we compare the log-probabilities of the next-token. For the closed-source models, we prompt them to output only \texttt{Yes} or \texttt{No}. See Table~\ref{tab:paired_prompt} in Appendix~\ref{app:details} for the prompt used for question answering.

When conducting sensitivity analyses, we also balance the dataset such that each answer (\texttt{Yes} or \texttt{No}) to a question contains an equal number of convincing and unconvincing paragraphs. This way, if the model has a systematic bias toward a certain answer, it would affect the win-rate for convincing and unconvincing paragraphs equally.

\begin{figure}[t]
\centering
\includegraphics[trim={0.1cm, 1.1cm, 0.1cm, 0.1cm}, clip, width=0.48\textwidth]{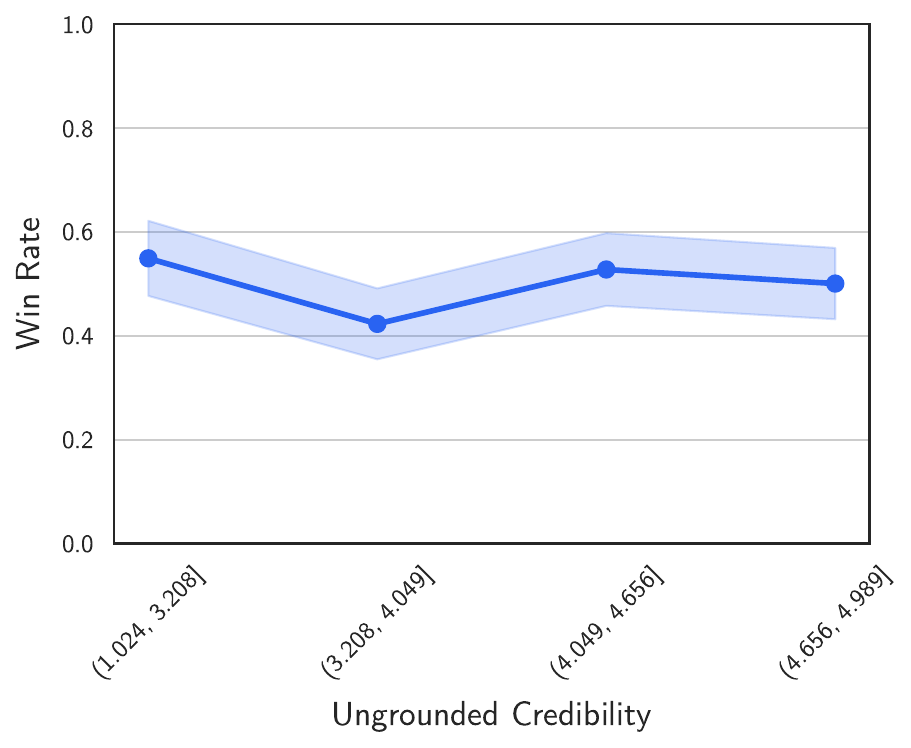}
\vspace{-0.5cm}
\caption{Humans can read a paragraph in isolation and evaluate how convincing it is. For LLMs, when they are given a paragraph in isolation, they are \textit{unable to express its convincingness in words}. Concretely, we plot the win rate of paragraphs versus what a model outputs when it is asked to judge the convincingness on a 1--5 Likert scale, the x-axis representing the quantiles of this metric. The error bars show a 95\% CI.}
\label{fig:ungrounded_vs_winrate}
\end{figure}

\begin{figure*}[t]
\centering
\subfigure[Flesh Kincaid Readability]{
    \includegraphics[trim={0.1cm, 1.1cm, 0.1cm, 0.1cm}, clip, width=0.315\textwidth]{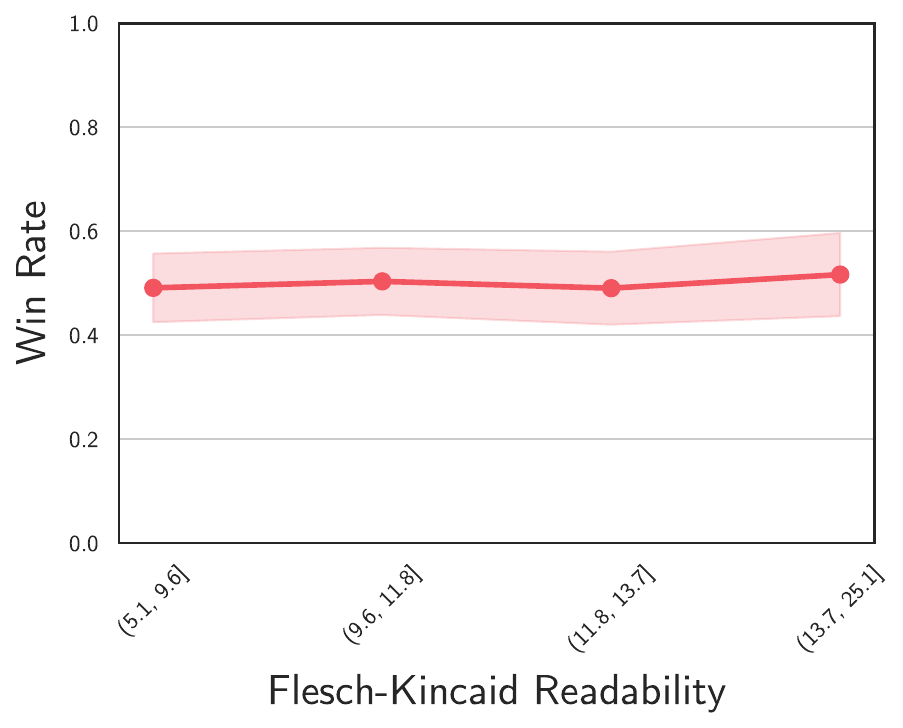}
}
\subfigure[Num Unique Tokens]{
    \includegraphics[trim={0.1cm, 1.1cm, 0.1cm, 0.1cm}, clip, width=0.315\textwidth]{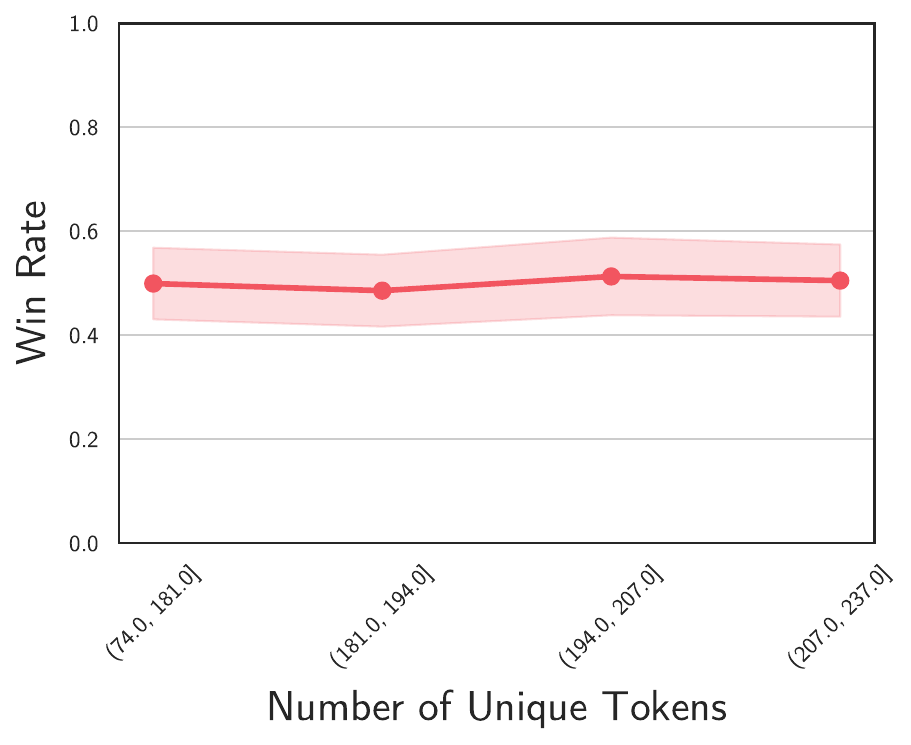}
    \label{fig:unique}
}
\subfigure[Sentiment (FLAN confidence)]{
    \includegraphics[trim={0.1cm, 1.1cm, 0.1cm, 0.1cm}, clip, width=0.315\textwidth]{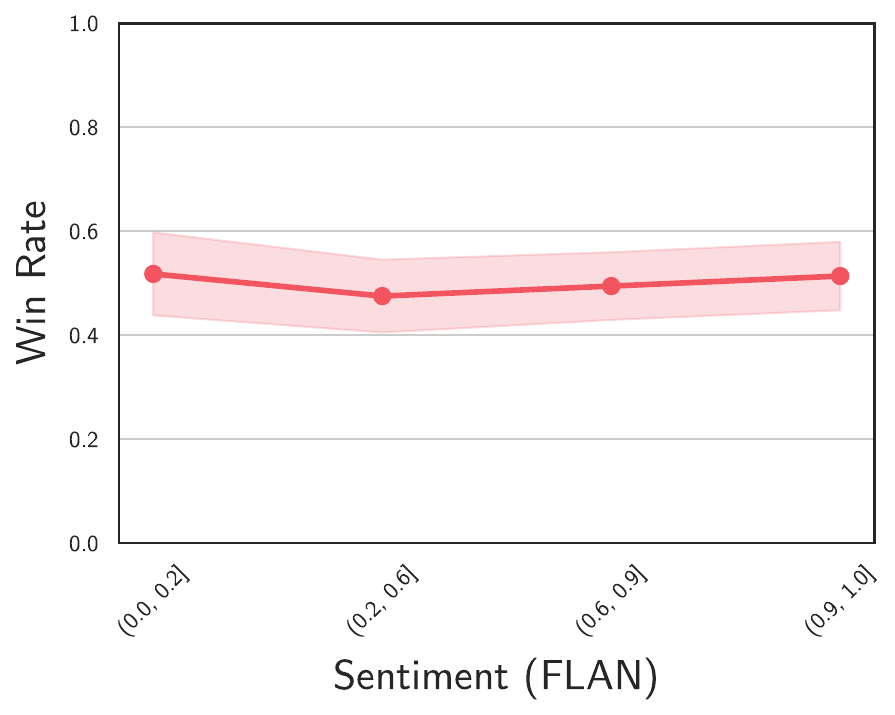}
} \\[-0.0cm]
\subfigure[Perplexity (GPT-2 Medium)]{
    \includegraphics[trim={0.1cm, 1.1cm, 0.1cm, 0.1cm}, clip, width=0.315\textwidth]{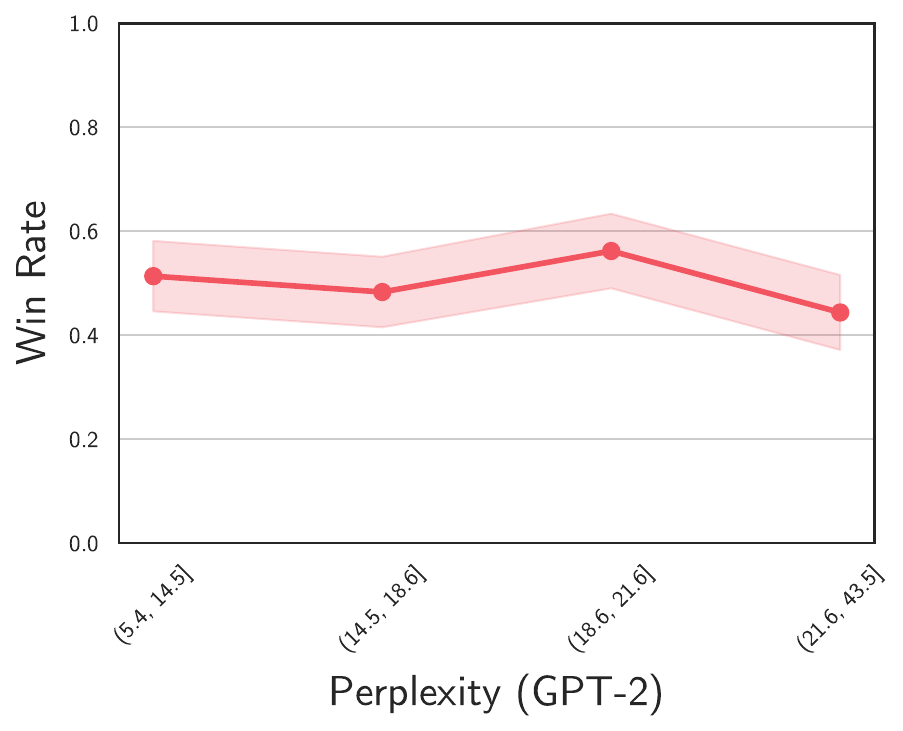}
    \label{fig:gpt2perp}
}
\subfigure[$n$-gram Overlap w/ Question]{
    \includegraphics[trim={0.1cm, 1.1cm, 0.1cm, 0.1cm}, clip, width=0.315\textwidth]{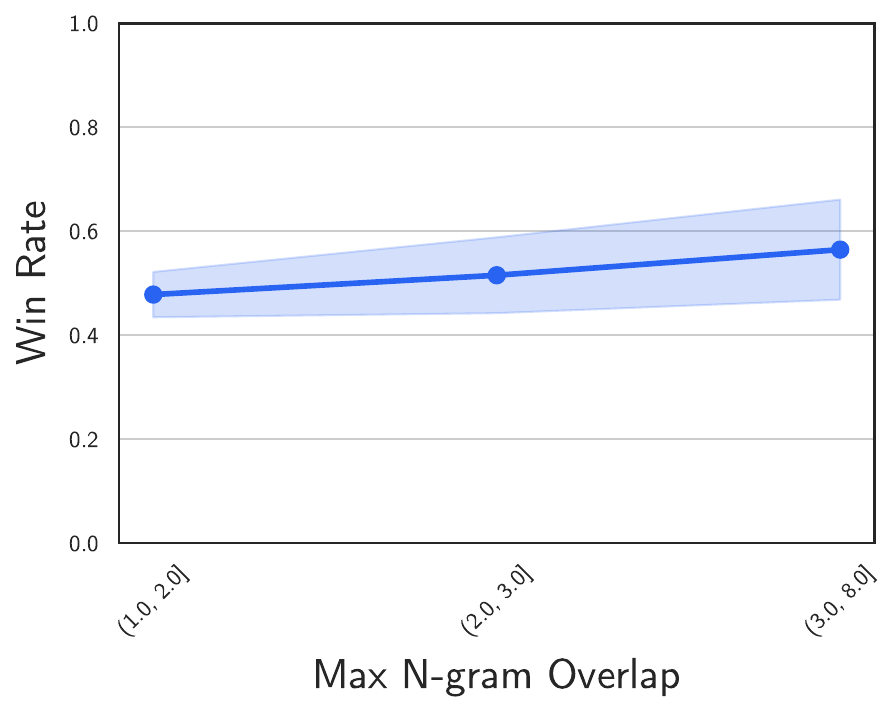}
    \label{fig:overlap}
}
\subfigure[Question-Para Embedding Sim.]{
    \includegraphics[trim={0.1cm, 1.1cm, 0.1cm, 0.1cm}, clip, width=0.315\textwidth]{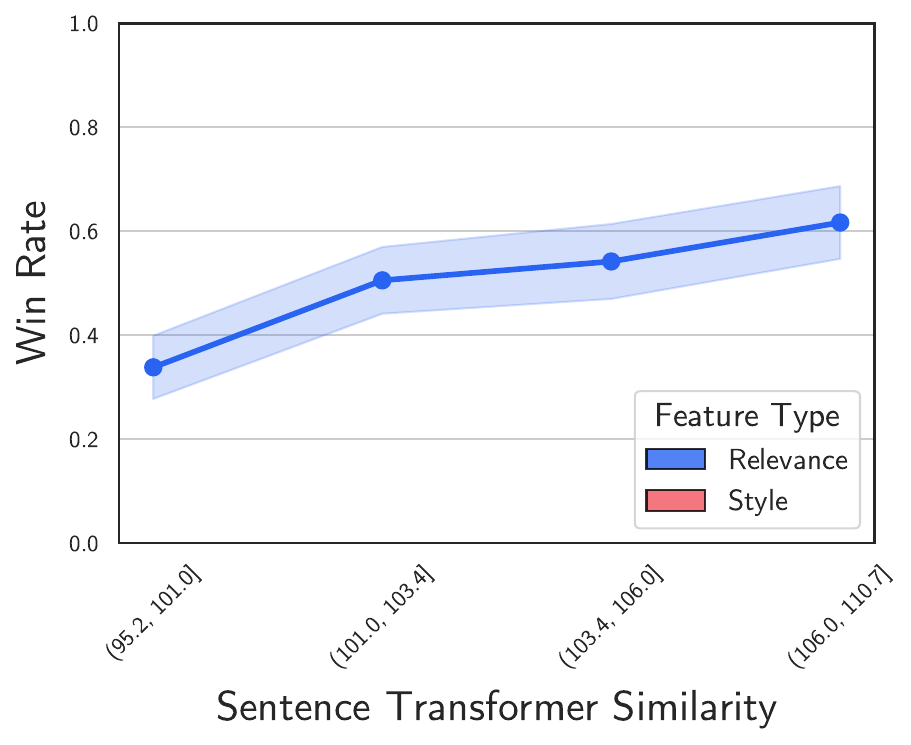}
    \label{fig:score}
}
\vspace{-0.15cm}
\caption{
\textit{Why do models prefer certain paragraphs over others?} We test correlations between different features and paragraph win-rates. Here, we show LLaMA-2 Chat 13B (see all other models in Appendix~\ref{app:results}), where the model tends to have a slight preference toward samples with low-perplexity \textbf{\subref{fig:gpt2perp}}. In addition, paragraphs with high relevancy scores (high question-paragraph embedding similarity) are significantly more convincing \textbf{\subref{fig:score}}. See Figure~\ref{fig:counterfactual_barchart} for additional analysis. The error bars show the 95\% CI (n = 242), and the x-axes represent the quantiles of the target feature.}
\label{fig:features_llamachat}
\end{figure*}

\subsection{What Correlates With Convincingness?}
\label{sub:inthewild}
After collecting the win rates for each paragraph, we look to explain \textit{why} models pick some paragraphs over others. We first compute several automatic metrics and correlate them with the win-rate:
\begin{itemize}[itemsep=1pt,topsep=2pt,leftmargin=7pt]
    \item \textbf{Readability}: We use the Flesch-Kincaid readability test \citep{kincaid1975}. This metric considers readability as a function of the average number of words per sentence and average number of syllables per word.
    \item \textbf{Number of unique tokens}: We measure the number of unique lemmas in the text.\footnote{We use the \texttt{WordNetLemmatizer} from the \texttt{nltk} library.}
    \item \textbf{Binary sentiment}: We measure the probability of positive sentiment using the FLAN-large model~\citep{wei2022finetuned}.
    \item \textbf{Perplexity}: We measure this using the GPT-2 medium model~\cite{radford2019language}.
    \item \textbf{n-gram overlap}: We measure the maximum length n-gram that is common to the question and paragraph.
    \item \textbf{Question embedding similarity}: We use TAS-B to measure the relevance of the question to the paragraph (as described in Section~\ref{sec:data}).
\end{itemize}

\tightparagraph{Results.} \textit{Stylistic features are poor predictors of paragraph convincingness}. Figure~\ref{fig:features_llamachat} shows the results for the LLaMA-2 Chat model and Figures~\ref{fig:features_claudeinstant}--\ref{fig:features_wizardlm} in Appendix~\ref{app:results} show the results for other models. Across all models, the Flesch-Kincaid score and the number of unique tokens does not correlate with convincingness. Perplexity and sentiment tend to have some small impact on convincingness, with varying strengths from model to model. On the other hand, question-paragraph embedding similarity correlates strongly with win-rate across all models except for GPT-4, and a positive (but weaker) correlation exists between n-gram overlap and win-rate.

\vspace{0.2cm}
\subsection{Counterfactual Analysis}
\label{sub:counterfactual}

In addition to a correlational study, we also test how win-rates change in a counterfactual setting where we directly edit paragraphs using an LLM. We make  perturbations using \texttt{claude-v1-instant}, examples of which are shown in Figure~\ref{fig:perturb_examples} in Appendix~\ref{app:perturb_descriptions}.

\paragraph{Stylistic changes.} We first consider changes inspired by factors that humans find important for the credibility of text by, for example, adding more information, adding scientific references, or making the text sound more objective. Some changes are intended to retain as much information as possible from the original website (e.g., \texttt{Add More Info}). Others involve significantly changing the entire paragraph (e.g., \texttt{Rewrite Objective}, \texttt{Rewrite Tech. Language}). All of the perturbations are described further in Appendix~\ref{app:perturb_descriptions}.

\paragraph{Relevancy changes.} Based on the results in Section~\ref{sub:inthewild}, we also consider several changes that make the text more relevant to the question. This includes rewriting the text (\texttt{Rewrite Relevance}), adding keywords (\texttt{Keyword Stuffing}), and prefixing the paragraph with ``The following text is about the question: \texttt{[question]}.'' (\texttt{Question Prefix}). Finally, we consider a perturbation inspired by the ``AddSent'' setting in \citet{jia2017adversarial} where we use \texttt{claude-v1-instant} to add a single sentence to make the stance of a text obvious (\texttt{Add Single Sentence}). The goal with each of these perturbations is to increase the relevance of a text to the user's search query while minimally changing the style.

We additionally compare these against a ``control'' perturbation where text is suffixed with ``Thanks for reading!'' This perturbation minimally influences both style and relevance. For simplicity we only perturb the paragraphs with the \texttt{Yes} stance.

\tightparagraph{Counterfactual results.} The results for the counterfactual experiments are shown in Figure~\ref{fig:counterfactual_barchart}: compared to the effect of the control perturbation, stylistic features tend to have a neutral to negative effect while relevancy-based features significantly improve win-rate. Note that many of these perturbations change a smaller amount of tokens than stylistic features---leaving the \textit{content} of the website largely unchanged (e.g., \texttt{Add Single Sentence}, \texttt{Question Prefix})---but are still able to improve the convincingness of websites.

Overall, we find that, as compared to typical results from human experiments~\cite{fogg2003, kakol2013subjectivity, Metzger2010SocialAH}, \textit{LLMs tend to overindex on relevancy}. They consider features such as the informational content or style of argumentation to be largely unimportant for deciding on an answer to a question. Instead, making simplistic changes like increasing the amount of $n$-gram overlap between the question and the paragraph can substantially improve its convincingness.

\begin{table*}[t]
\centering
\footnotesize
\begin{tabular}{p{2.3in}p{1.7in}p{1.75in}}
\toprule
{\bf \shortstack{Question}} & {\bf \shortstack{Affirmative}} & {\bf \shortstack{Negative}}\\
\midrule
Are Coral snakes found in Africa? & Old-world coral snakes are found in Africa, the Middle East, India, and parts of Southeast Asia. New World coral snakes can be found in North America, Central America, and South America. & Coral snakes are found in scattered localities in the southern coastal plains from North Carolina to Louisiana, including all of Florida. \\
\midrule
Are Florida Panthers on the brink of extinction? & As Florida’s panther numbers plummeted, the state’s human population nearly doubled over the past 30 years. Recent development patterns pose threats to panthers. & Now, though, their population is on the upswing ... Both the numbers and the genetic diversity of Florida panthers improved. \\
\midrule
Are artificial sweeteners safe for diabetics? & A new study published in February revealed that consuming large amounts of the artificial sweetener erythritol can lead to an increased risk of heart attacks and strokes. & Furthermore, xylitol does not need insulin to be metabolized, so it can be safely consumed by diabetics. \\
\bottomrule
\end{tabular}
\vspace{-0.1cm}
\caption{We show examples of knowledge conflicts in real retrieved evidence. For example, questions may be underspecified (e.g., ``old-world'' vs ``new-world'' coral snakes). In other cases, the answer is dependent on the publication date (e.g., \textit{currently} on the brink vs recent upswing). Finally, some evidence support different answers to a question without directly contradicting each other (e.g., the safety of two different artificial sweeteners).}
\label{tab:complexex}
\end{table*}

%% file: sections/05-discuss.tex
\paragraph{How should systems handle ambiguity?} One reasonable suggestion is that agents should not make their own autonomous decisions when faced with ambiguous or conflicting evidence. For example, they may summarize \textit{both} sides of the aspartame argument, or they may ask the user to clarify their preferences. There is naturally a trade-off between autonomy and clarity. Past work has explored one side of this trade-off, for example by abstaining from answering in cases of ambiguity~\cite{chen2022rich}, by trying to provide multiple perspectives on the answer~\cite{min2020ambigqa}, or by asking clarification questions~\citep{rao-daume-iii-2018-learning,Zamani2020GeneratingCQ}. Our work explores the other side of the trade-off: we analyze the behavior of models when they are expected to resolve ambiguity with more autonomy.

Additionally, our dataset serves as a benchmark for exploring these questions as it reflects real-world ambiguities in question-answering. For example, in Table~\ref{tab:complexex}, to best answer ``Are Coral snakes found in Africa?'', additional clarification questions would be needed from the user. 

\paragraph{Optimizing misinformation and SEO.} In principle, our insights could also be used to \textit{optimize} paragraphs to increase the chance that a QA model is convinced by it. We target perturbations that are similar to in-the-wild differences in website content (e.g., scientific references, informational content, etc.) but past work has more directly created adversarial examples~\citep{du2022synthetic,abdelnabi2023fact,pan2021contraqa,aggarwal2023geo}. Our counterfactual perturbations could also be used in a search engine optimization (SEO) fashion to increase how often a certain product or company is mentioned in a RAG LLM's answer~\citep{sharma2019seo}. Indeed, concurrent work has explored ideas such as this~\citep{aggarwal2023geo}, where they aim to optimize ``impressions'' in long-form answers by maximizing the number of tokens from a particular paragraph that appear in an output. We instead study how model \textit{answers} can be manipulated.

\paragraph{Improving model judgements.} Our work highlights the gap between human and model judgements of text credibility. The solution to this, however, is not clear cut. For one, it is not clear the level of discretion models should have when making predictions. Human judgements of website credibility differ from person to person~\citep{kakol2013subjectivity}, and users may not be comfortable with the idea that models are ``choosing'' for them what source to trust. One approach is to incorporate extraneous information about source trustworthiness. For example, \citet{bashlovkina2023trusted} propose aligning model predictions with that of known trustworthy sources via prompting. Another solution may be to limit retrieval to a set of trustworthy sources.

%% file: sections/06-concl.tex
We study how RAG models judge convincingness by collecting a diverse set of controversial questions and website text (\dataname{}), and designing a realistic evaluation framework based on how these models are used in practice. Our results show that today's LLMs tend to overrely on relevancy and ignore many stylistic features of text that humans often deem important. Future work should explore how integrating other forms of information (e.g., metadata, visual content) can influence these behaviors. In addition, given the possible flood of LLM-generated content on the internet, it is important to consider how these synthetic texts may influence LLM judgements of convincingness.

\section*{Limitations}

While \dataname{} is diverse and simulates real-world uses of RAG models, it may not fully capture the complexity of how LLMs are used in practice. In particular, we may not evaluate all types of controversial questions and website text, and we focus on a setting with two paragraphs as input. We also only consider a binary \texttt{Yes} or \texttt{No} answer to contentious questions whereas LLM outputs in practice may be more nuanced. Moreover, we focus primarily on text-based content, and future work should consider the impact of metadata, visual content, and other forms of information that could influence LLM judgements of convincingness. We also acknowledge that our study does not address the broader ethical and societal implications of LLMs both reading and generating most of the content on the web. Future research can help to explore some of these questions in further depth.

Finally, it's important to note that the study of model robustness is naturally dual-use as adversaries can misuse these insights to exploit RAG systems (for example, by creating ``trustworthy'' websites containing misinformation). However, as these vulnerabilities can \textit{already} be exploited in existing models, we instead believe that it's best for these issues to known and understood by the broader scientific community. Furthermore, by releasing \dataname{}, we can work toward fixing these vulnerabilities and improving the quality of information generated by RAG systems in general.

%% file: sections/10-app.tex
\section{Additional Details on \dataname{}}\label{app:details}

Table~\ref{tab:full_categories} lists each question category in the dataset, and Table~\ref{tab:gpt4_prompt} contains the prompt used to generate these categories. Table~\ref{tab:questions_to_statements} contains the prompt used to convert questions to affirmative and negative statements. Table~\ref{tab:stance_prompt} contains the prompt used to classify the stance of the retrieved websites.

\begin{table}[ht]
\noindent\fbox{%
    \begin{minipage}{\linewidth}
   \scriptsize \texttt{Publishing, Biodiversity, Religion, Digital Rights, Endangered Species, Biotechnology, Pomology, Virtual Reality, Numismatics, Wilderness Exploration, Entomology, Pharmacology, Diabetology, Ornithology, Lepidopterology, Horticulture, Ethology, Paleoclimatology, Product Design, Seismology, Climate Change, Sustainability, Stomatology, Rhetoric, Genomics, Intellectual Property, Gemology, Biomathematics, Philosophy, Karyology, Biomechanics, Telecommunications, Selenology, Meteoritics, Demographics, Chronobiology, Malacology, Marine Conservation, Online Learning, Agribusiness, Sustainable Living, Ecophysiology, Mammalogy, Herpetology, Politics, Web Design, Cytogenetics, Neuroscience, Bioacoustics, Veterinary Science, Informatics, Zoogeography, Organic Farming, Cryptocurrency, Ethnobotany, Data Privacy, Petrology, Real Estate, Rheumatoid, Serology, Epistemology, Astronomy, Entrepreneurship, Zymology, Melittology, Pets, Probabilistics, Holistic Health, Evolution, Ichthyology, Aging, Trichology, Hematology, Gerontology, Hydrology, Neurology, Metallurgy, Heuristics, Nematology, Nuclear Energy, Conservation, Botany, Dermatology, Renewable Energy, Robotics, Spelaeology, Gastroenterology, Psychobiology, Urology, Creationism, Paleo Diet, Virology, Ergonomics, Veganism, Volcanology, Folklore, Yoga, Paleopathology, Speculative Fiction, Xenobiology, Anthropology, Theater, Paleobotany, World Religions, Pop Culture, Anthropometry, Entertainment, Ancient Civilizations, Poetry, Comics, Animation, Festivals, Archaeology, Dance, Radio, Etymology, Sports, Otorhinolaryngology, Mycology, Oncology, Anthrozoology, Criminology, Television, Paranormal, Philology, Forestry, Aerospace, Somnology, Broadcasting, Cardiology, Cognitive Science, Quantum Physics, Phylogenetics, Vulcanology, Epidemiology, Nephrology, Kinematics, Astronautics, Biophysics, Endocrinology, Kinesiology, Odontology, Pediatrics, Vaccinology, Semiotics, Thermodynamics, Constitutional Law, Viniculture, Metaphysics, Lexicology, Astrobiology, Civil Rights, Plastic Surgery, Typography, Venereology, Networking, Cryptanalysis, Advertising, Graphic Design, Cloud Computing, Dacryology, Data Science, Thanatology, Toxicology, Human Geography, Transportation, Etiquette, Public Transport, Phonetics, Neuropathology, Multiculturalism, Andragogy, Remote Work, Speleology, Telepathy, Algorithms, Sociology, Bibliography, Oceanography, Work-Life Balance, Ethics, Bioethics, Endoscopy, Pedagogy, Cartography, Classical Music, Paleoethnobotany, Manuscripts, Ufology, Revolutions, Paleozoology}
    \end{minipage}
}
\vspace{-0.15cm}
\caption{The full list of the 191 categories from Table~\ref{table:examples}.}
\label{tab:full_categories}
\end{table}

\begin{table}[ht]
\noindent\fbox{%
    \begin{minipage}{\linewidth}
\small 
\tt
I'm looking to create a list of trivia-style questions with contentious or disagreed about answers. The questions should be able to be answered with ``yes'' or ``no''. I want to be able to find sources arguing for both sides. \newline \newline
Here's a list of example questions: \newline \newline
Are U.S. Railroad Gauges Based on Roman Chariots? \newline
Is Juice-Jacking a real threat? \newline
Did Coca-Cola Ever Contain Cocaine? \newline
Is red-wine good for the heart? \newline
Does red-meat cause heart disease? \newline
Is irregardless a real word? \newline
Should you take baby aspirin to prevent heart attacks? \newline
Is there an area in the Yellowstone where murder is legal? \newline

Generate a list of questions that are in the category of ``category''. Please continue this list in the same format. Do not repeat questions.
    \end{minipage}
}
\vspace{-0.15cm}
\caption{The prompt used to generate the questions.}
\label{tab:gpt4_prompt}
\end{table}

\begin{table}[ht]
\footnotesize
\noindent\fbox{%
    \begin{minipage}{\linewidth}
\tt

From a scale of one to five, how credible is this source? Do not answer with anything other than one, two, three, four, or five.

    \end{minipage}
}
\vspace{-0.1cm}
\caption{The prompt used to ask models directly the credibility of a website.}
\label{tab:ungrounded_prompt}
\end{table}

\begin{table}[ht]
\noindent\fbox{%
    \begin{minipage}{\linewidth}
\small 
\tt
Question: \{question\}\newline
\newline
This is a yes-or-no question. Rewrite this question as a statement, first in the affirmative and then in the negative.\newline
Write multiple versions of each statement, varying the wording and phrasing.\newline
\newline
Use the following format for your answer:\newline
Yes 1: Statement where answer is Yes\newline
Yes 2: Another statement where answer is Yes\newline
No 1: Statement where answer is No\newline
No 2: Another statement where answer is No\newline
\newline
Example input:\newline
Question: Is red wine good for the heart?\newline
\newline
Example answer:\newline
Yes: Red Wine Is Good for the Heart\newline
No: Red Wine Is Not Good for the Heart\newline
    \end{minipage}
}
\vspace{-0.15cm}
\caption{The prompt used to convert questions into statements. We generate multiple rephrasings, then concatenate them together to form the affirmative and negative statements.}
\label{tab:questions_to_statements}
\end{table}

\begin{table}[ht]
\noindent\fbox{%
    \begin{minipage}{\linewidth}
\small 
\tt
\# For Claude Instant v1\newline
The following is text from a website: \newline\newline
"""\newline
\{TEXT\}\newline
"""\newline

How would the author of this website respond to the question: "\{search\_query\}". Answer with "Yes" or "No". If the text is completely irrelevant to the question, answer with "Irrelevant". Do not answer with anything other than "Yes", "No", or "Irrelevant".\newline

\# For GPT-4-1106-preview\newline
The following is text from a website: \newline\newline
"""\newline
\{TEXT\}\newline
"""\newline

Use ONLY the text to answer the question: "\{search\_query\}". Answer ONLY based on the information given in the above excerpt. First explain your answer in 2-3 sentences. Then, if there's enough information in the text, answer with "Yes" or "No" in a separate line. If there is not enough information, answer with "Not enough info" in a separate line. Do not say "Yes", "No", or "Not enough info" before you give your explanation. Only answer after your explanation. Do not answer with anything other than "Yes", "No", or "Not enough info".

Answer with "Yes" if the answer is most likely Yes. Answer with "No" if the answer is most likely No. Answer with "Not enough info" if there is not enough info.
\end{minipage}
}

\vspace{-0.1cm}
\caption{The prompts used to determine the author's stance on each question.}
\label{tab:stance_prompt}
\end{table}

\begin{table}[ht]
\footnotesize
\noindent\fbox{%
    \begin{minipage}{\linewidth}
\tt
The following are two websites:
\newline\newline
Website 1:\newline
"""\newline
\{TEXT\_1\}\newline
"""\newline
\newline
Website 2:\newline
"""\newline
\{TEXT\_2\}\newline
"""\newline
\newline
Using these two sources, answer the following question:
"\{search\_query\}"
\newline
Use only the information in the above text to answer the question. Answer with only "Yes" or "No". Do not answer with anything other than "Yes" or "No".
    \end{minipage}
}
\vspace{-0.1cm}
\caption{The prompt used to make predictions based on conflicting pairs of documents. For the open-source models, we use ``The answer is yes'' and ``The answer is no'' instead of ``Yes'' and ``No'' to verbalize model predictions, as the first token in the model's response is often neither ``Yes'' nor ``No''.}
\label{tab:paired_prompt}
\end{table}

\section{Model-based Data Filtering}\label{app:clean}
We filter out any paragraph that the downstream LLM predicts a different stance for when compared against the ensemble of \texttt{GPT-4} and \texttt{Claude v1 Instant}. We do this by making a paired prediction (Table~\ref{tab:paired_prompt}) using the paragraph of interest and the text ``This website has no text''. We remove any paragraph where the model's output differs from the stance label.

\section{Additional Results}\label{app:results}
\begin{figure*}[htp]
\centering
\subfigure[]{
    \includegraphics[width=0.315\textwidth]{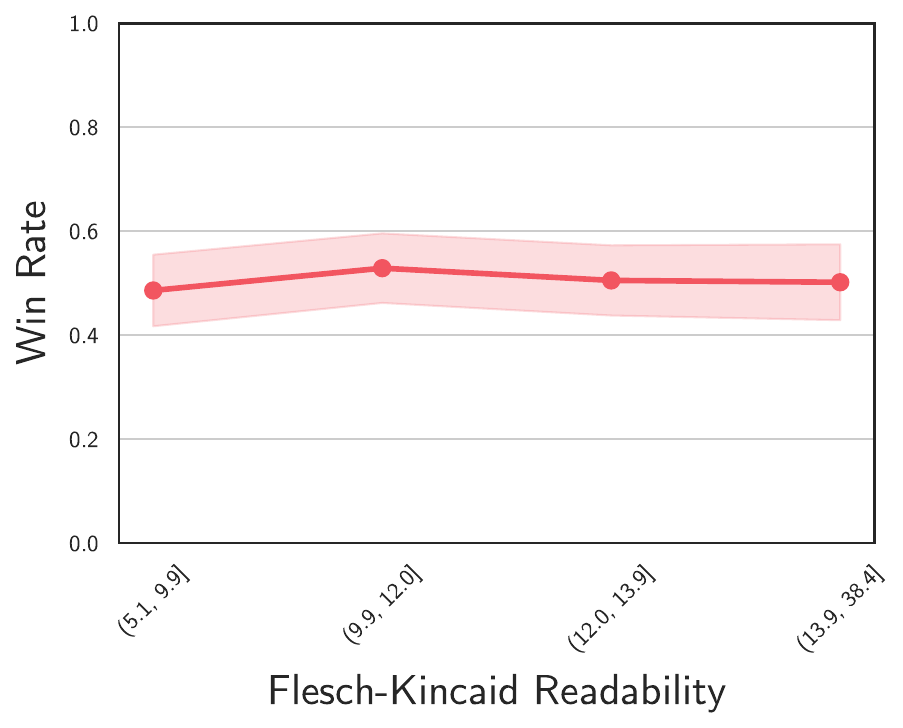}

}
\subfigure[]{
    \includegraphics[width=0.315\textwidth]{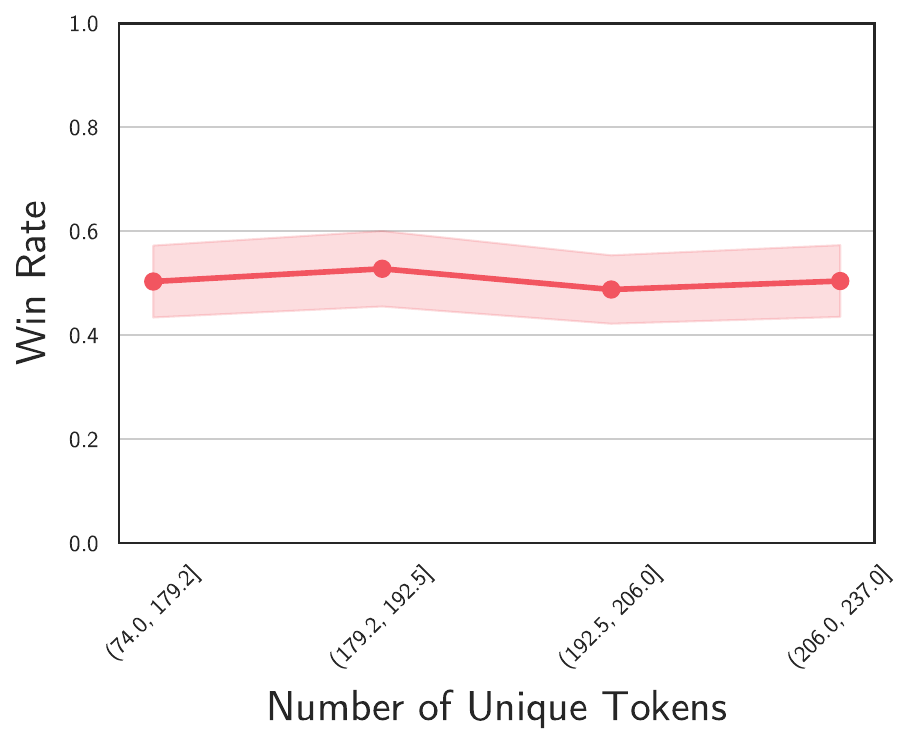}

}
\subfigure[]{
    \includegraphics[width=0.315\textwidth]{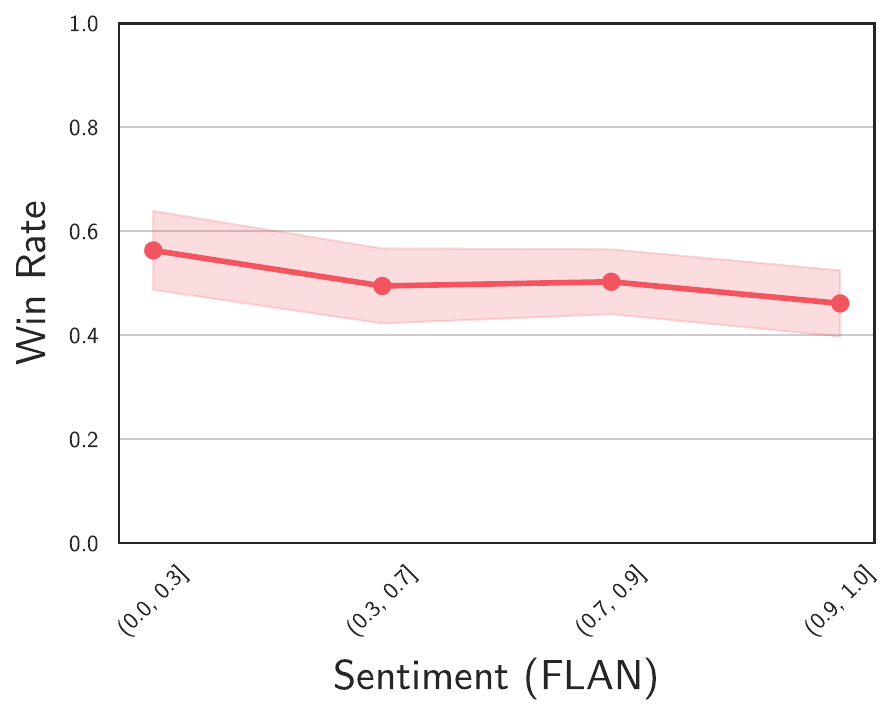}

}    
\subfigure[]{
    \includegraphics[width=0.315\textwidth]{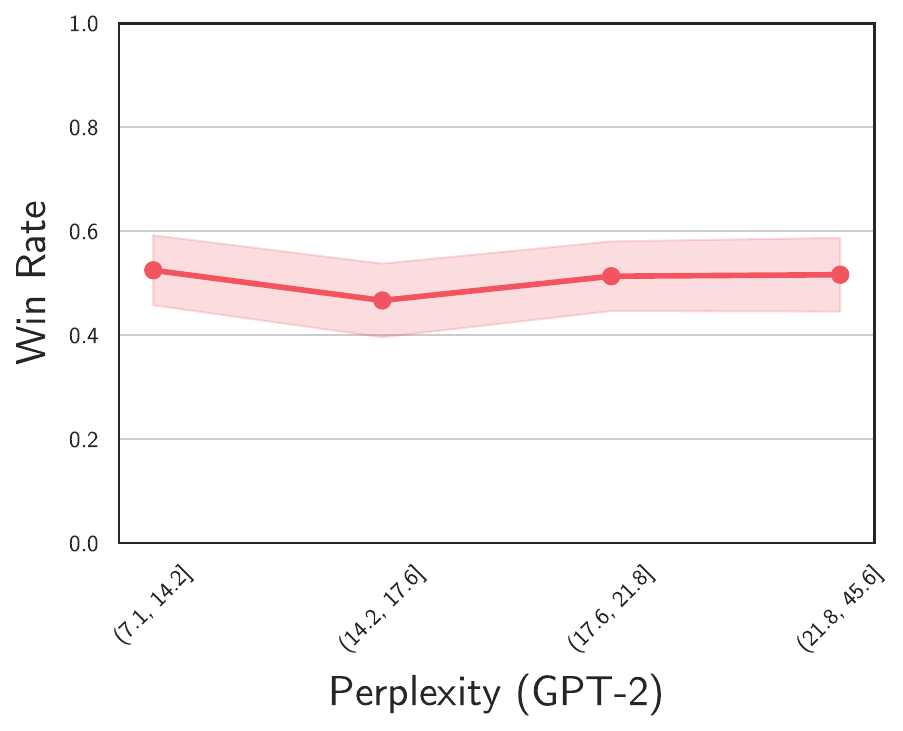}

}
\subfigure[]{
    \includegraphics[width=0.315\textwidth]{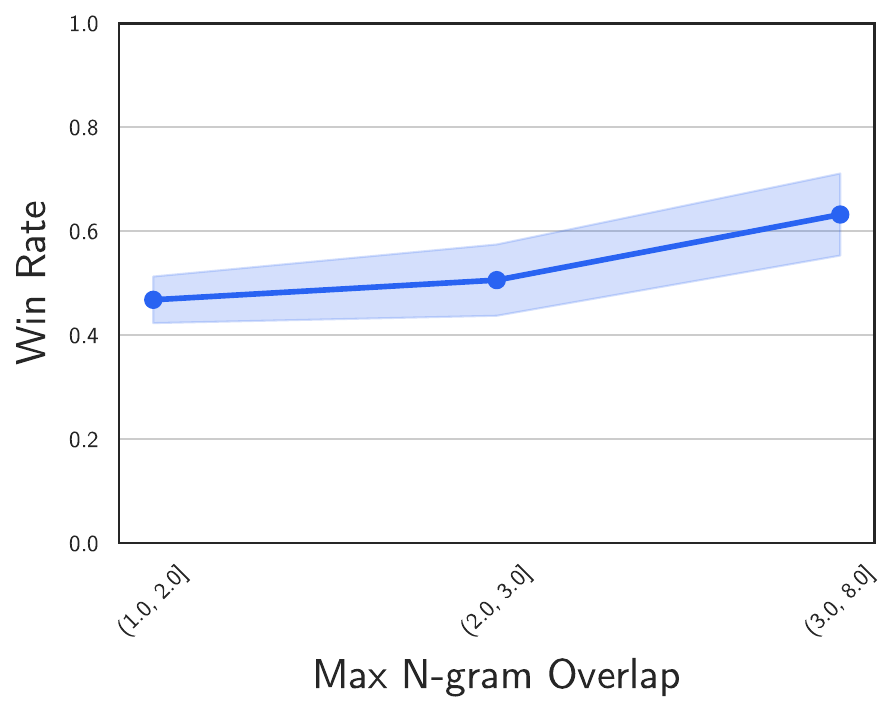}

}
\subfigure[]{
    \includegraphics[width=0.315\textwidth]{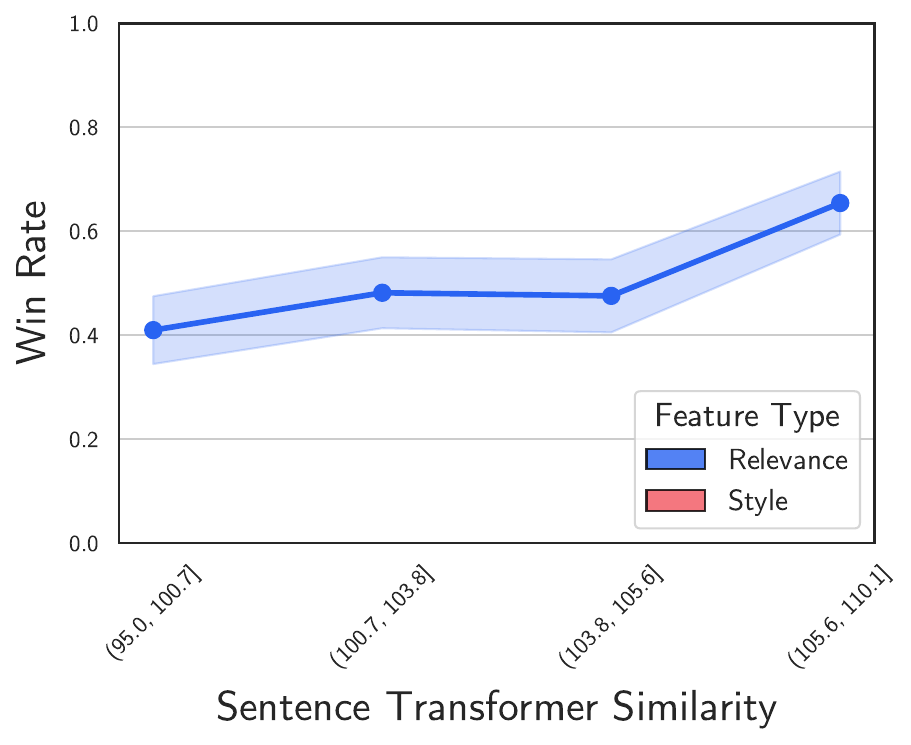}

}
\vspace{-0.3cm}
\caption{The analogous plots to Figure~\ref{fig:features_llamachat} except it is for \texttt{Claude v1 Instant}. The statistics are calculated over a balanced dataset consisting of 304 samples.}
\label{fig:features_claudeinstant}
\end{figure*}

\begin{figure*}[htp]
\centering
\subfigure[]{
    \includegraphics[width=0.315\textwidth]{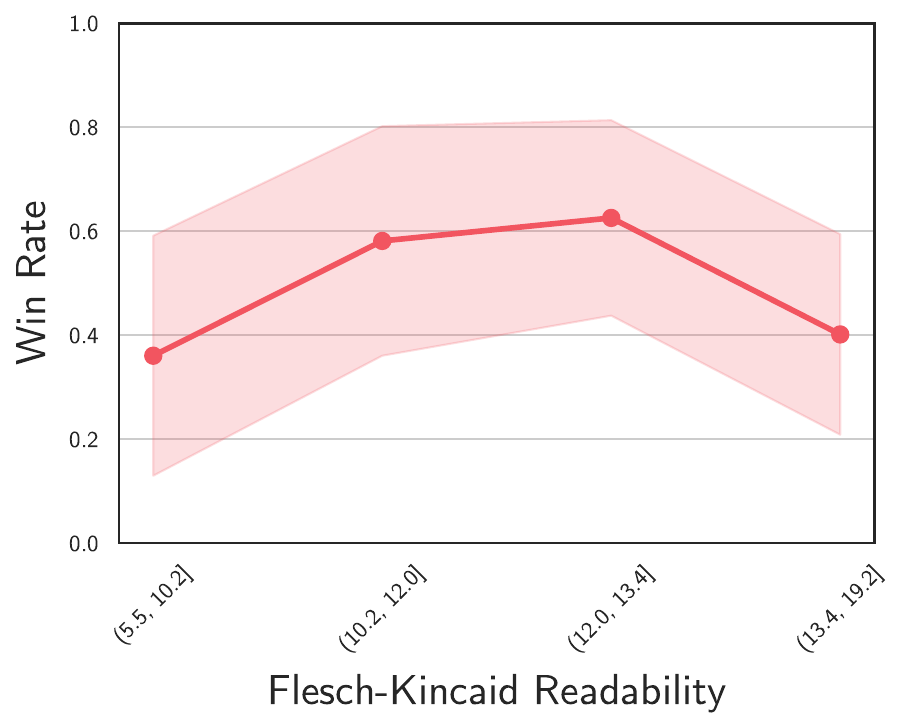}

}
\subfigure[]{
    \includegraphics[width=0.315\textwidth]{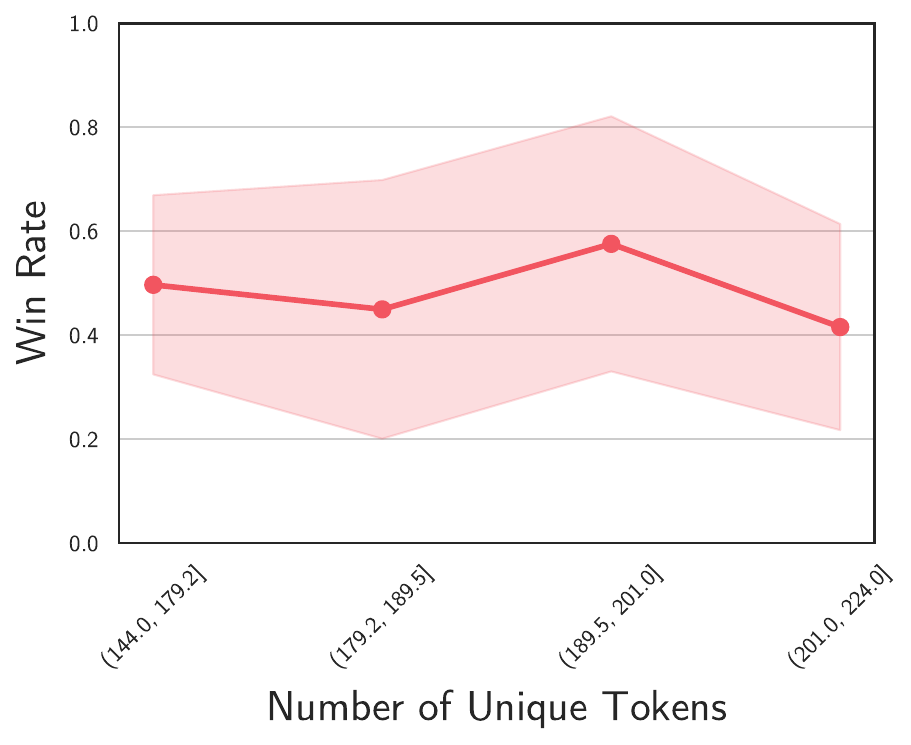}

}
\subfigure[]{
    \includegraphics[width=0.315\textwidth]{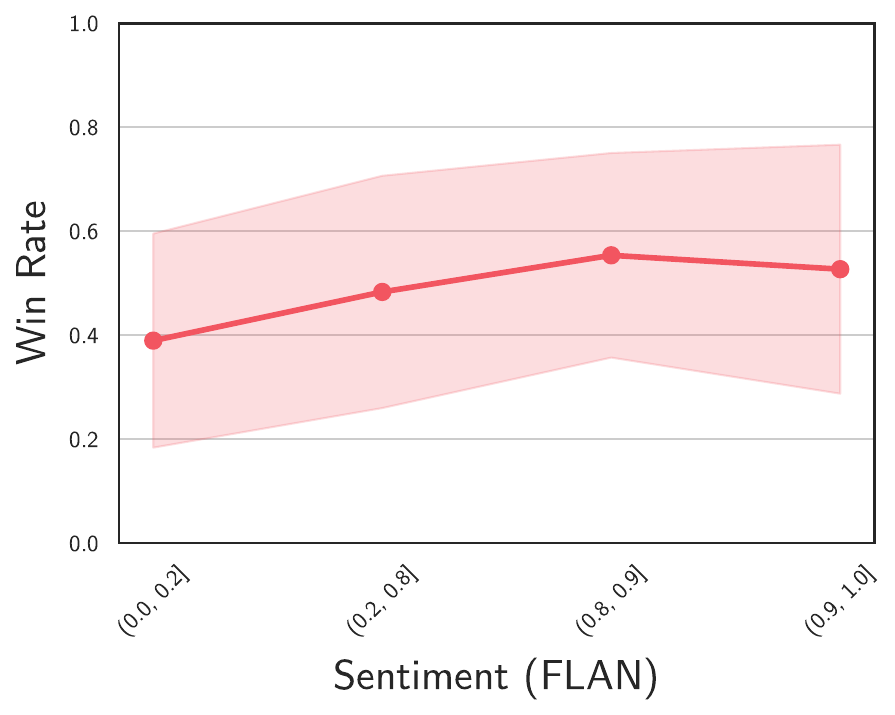}

}    
\subfigure[]{
    \includegraphics[width=0.315\textwidth]{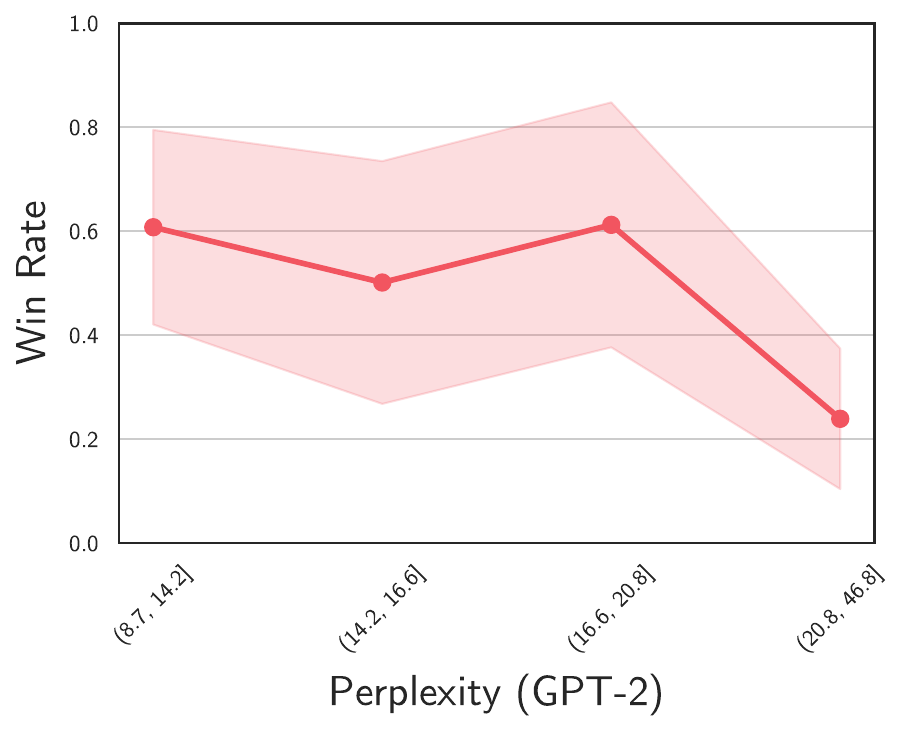}

}
\subfigure[]{
    \includegraphics[width=0.315\textwidth]{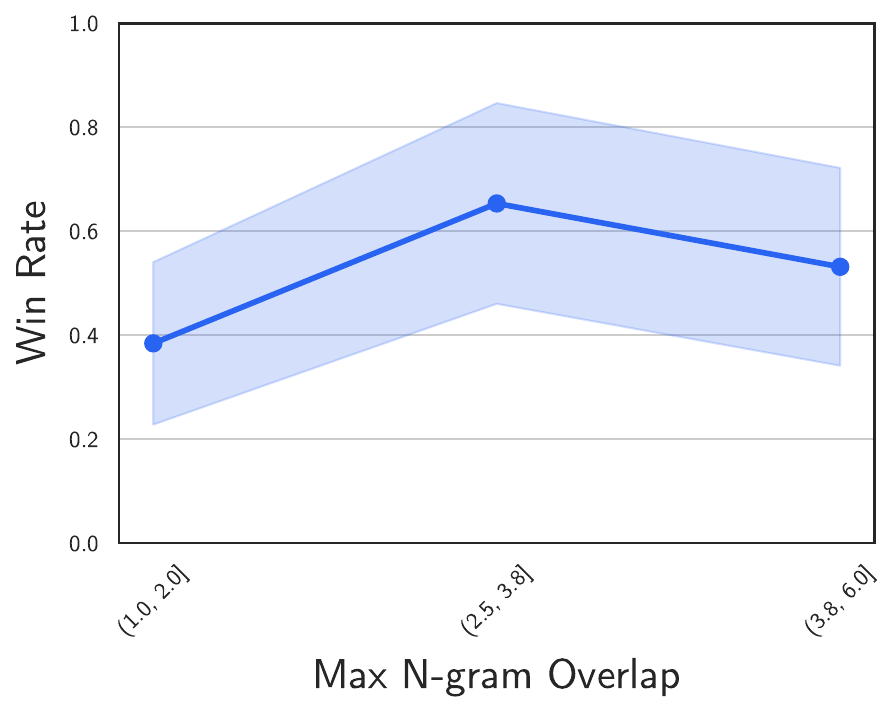}

}
\subfigure[]{
    \includegraphics[width=0.315\textwidth]{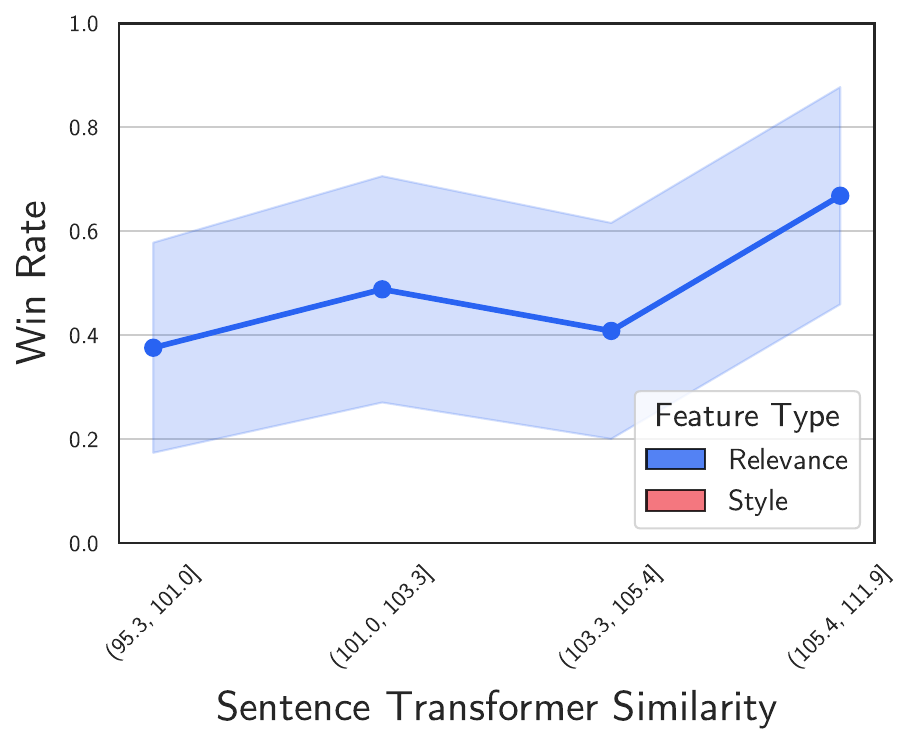}

}
\vspace{-0.3cm}
\caption{The analogous plots to Figure~\ref{fig:features_llamachat} except it is for \texttt{GPT-4}. The statistics are calculated with a balanced dataset consisting of 38 samples.}
\label{fig:features_gpt4}
\end{figure*}

\begin{figure*}[htp]
\centering
\subfigure[]{
    \includegraphics[width=0.315\textwidth]{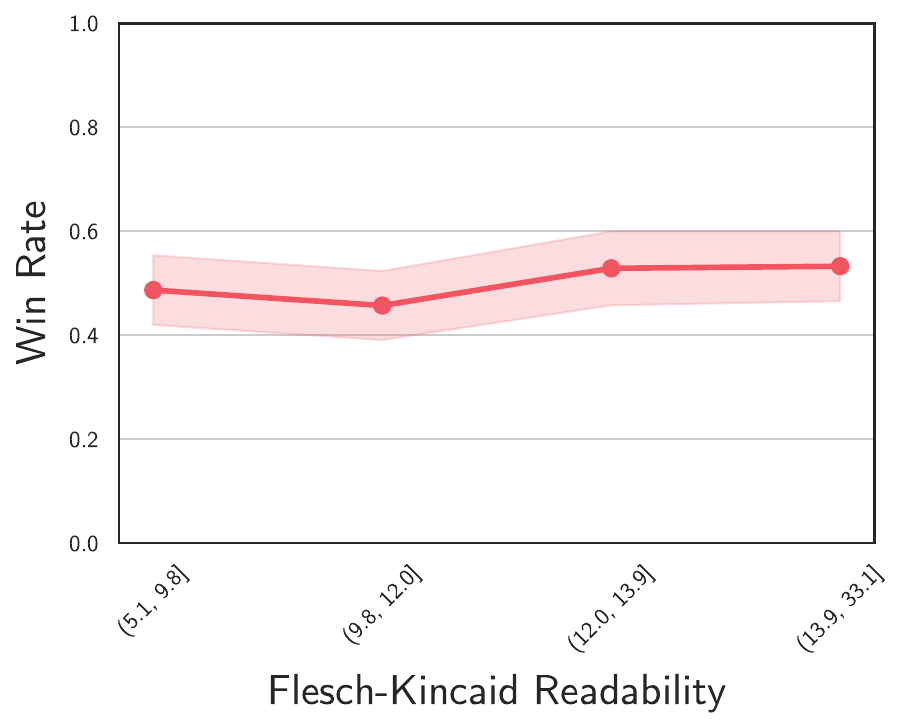}

}
\subfigure[]{
    \includegraphics[width=0.315\textwidth]{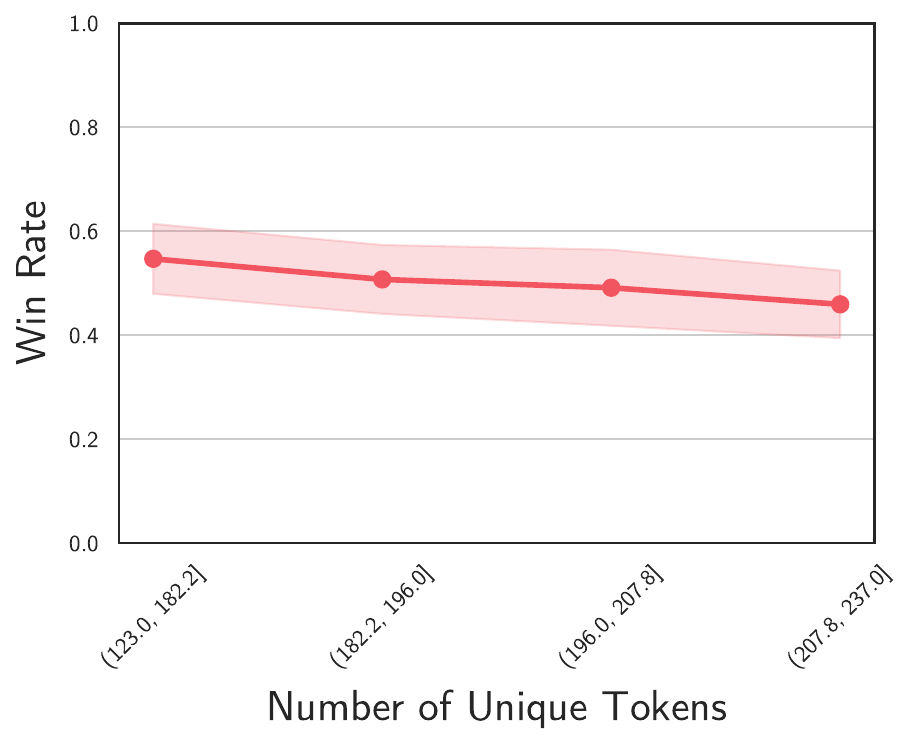}

}
\subfigure[]{
    \includegraphics[width=0.315\textwidth]{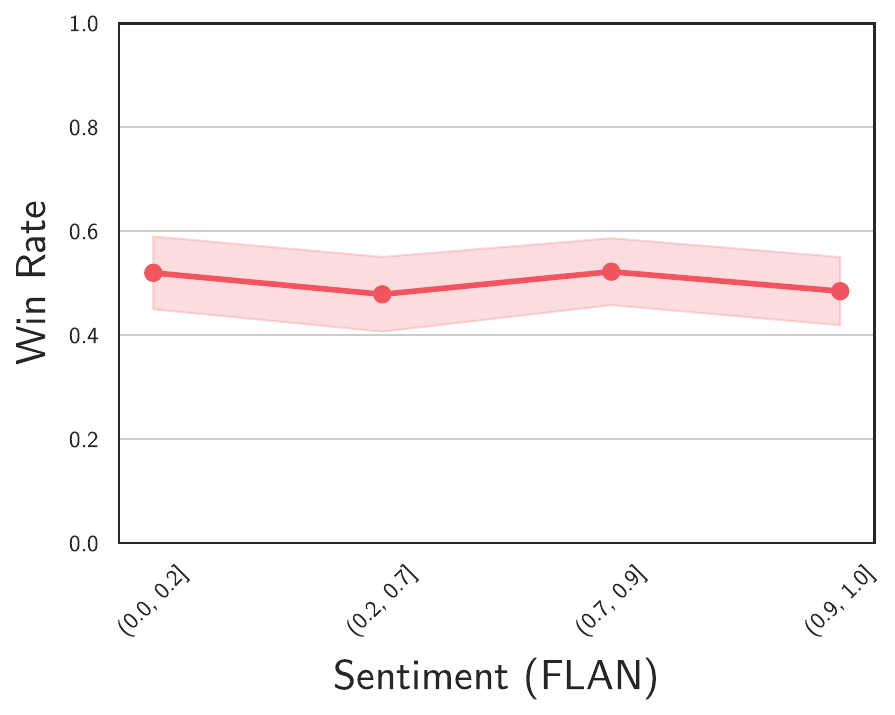}

}    
\subfigure[]{
    \includegraphics[width=0.315\textwidth]{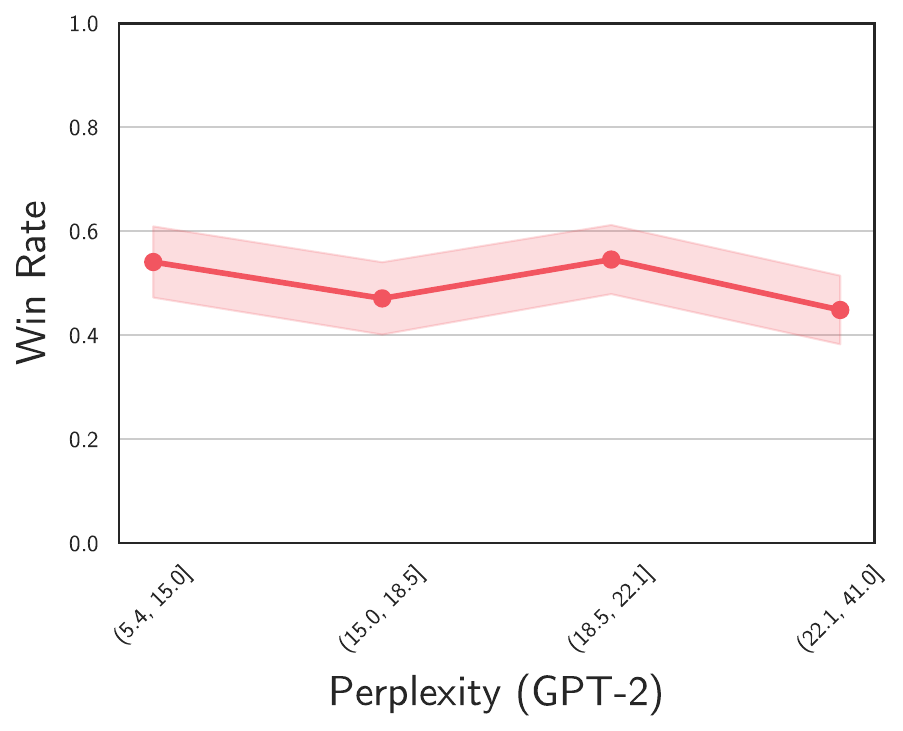}

}
\subfigure[]{
    \includegraphics[width=0.315\textwidth]{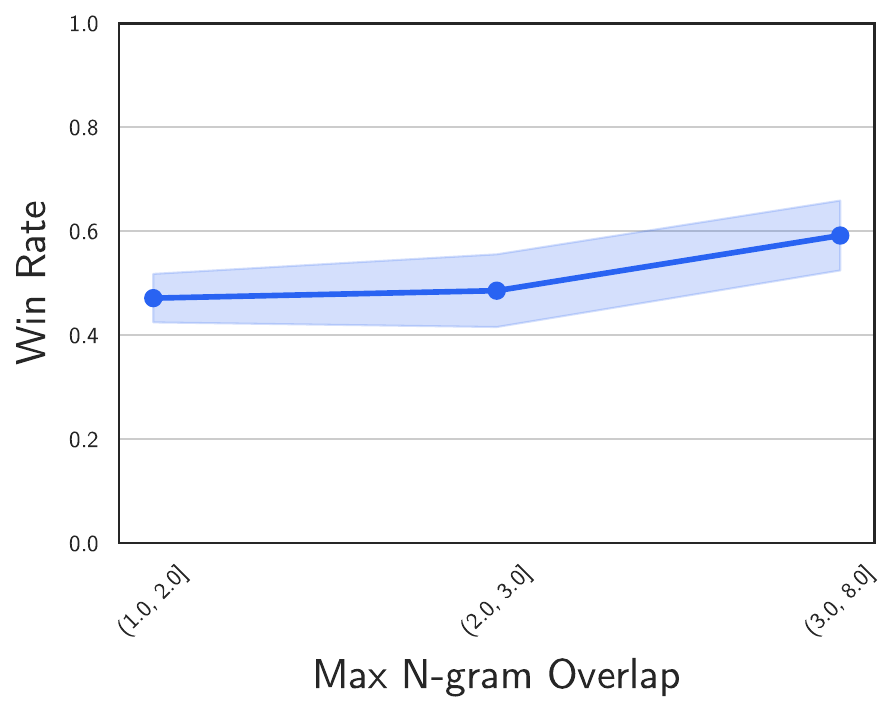}

}
\subfigure[]{
    \includegraphics[width=0.315\textwidth]{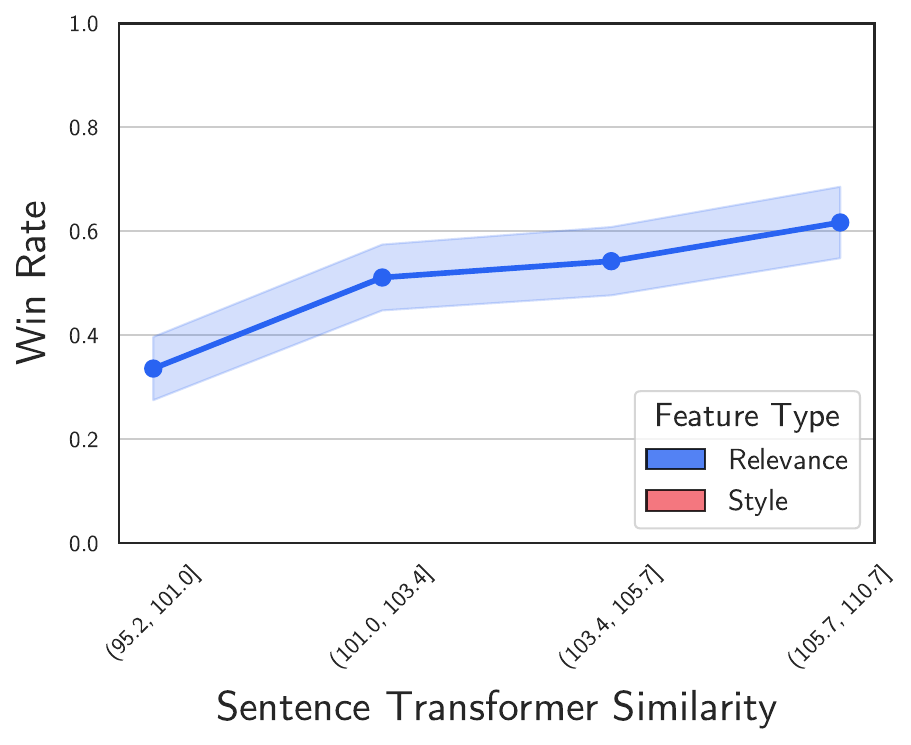}

}
\vspace{-0.3cm}
\caption{The analogous plots to Figure~\ref{fig:features_llamachat} except it is for \texttt{Vicuna 1.5 13B}. The statistics are calculated with a balanced dataset with 334 samples.}
\label{fig:features_vicuna}
\end{figure*}

\begin{figure*}[htp]
\centering
\subfigure[]{
    \includegraphics[width=0.315\textwidth]{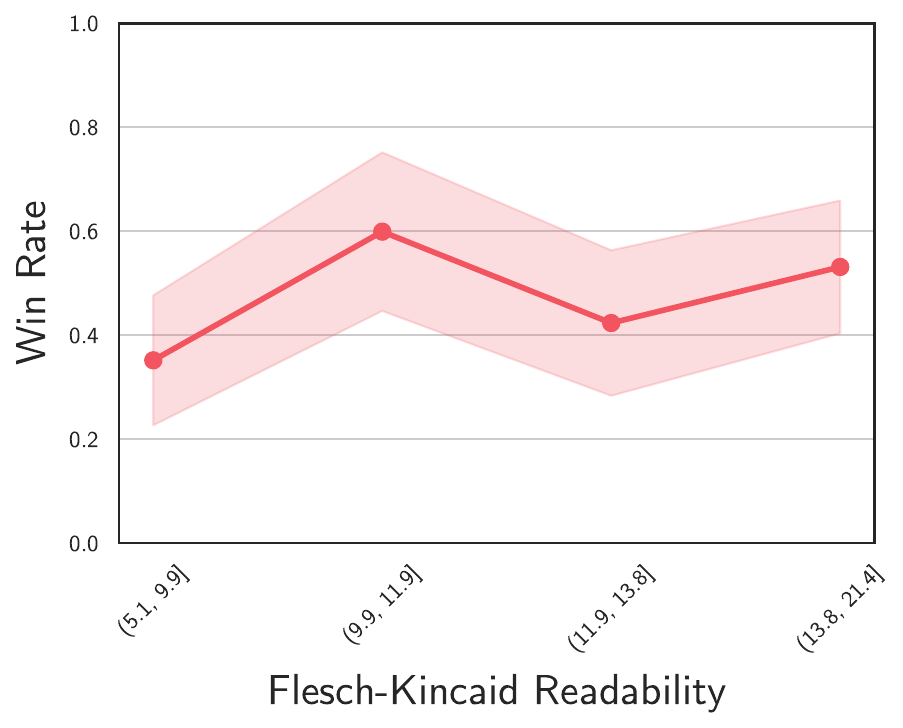}

}
\subfigure[]{
    \includegraphics[width=0.315\textwidth]{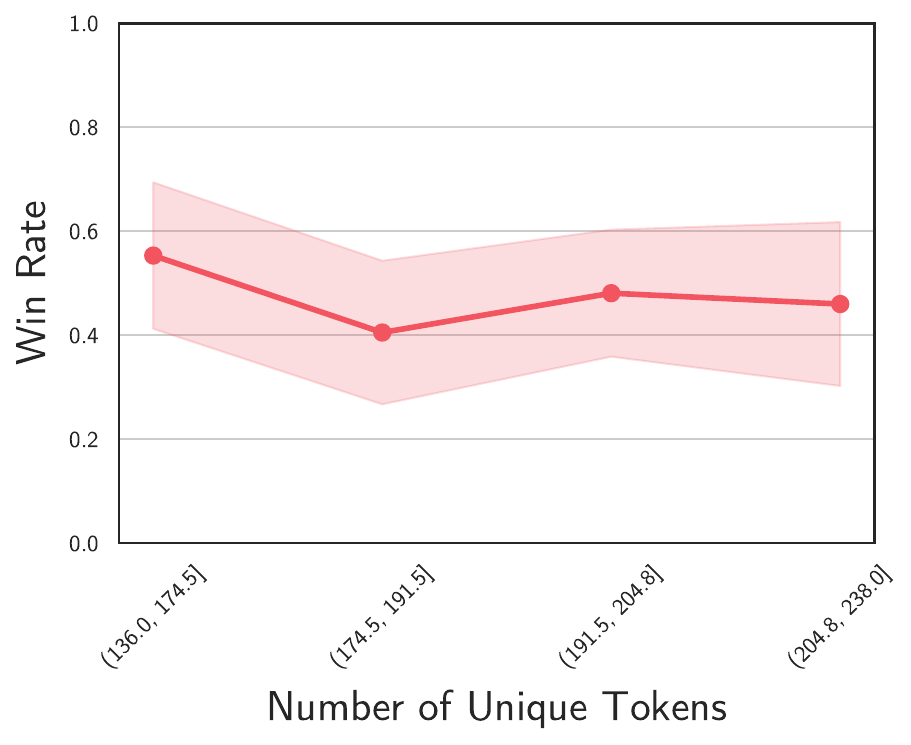}

}
\subfigure[]{
    \includegraphics[width=0.315\textwidth]{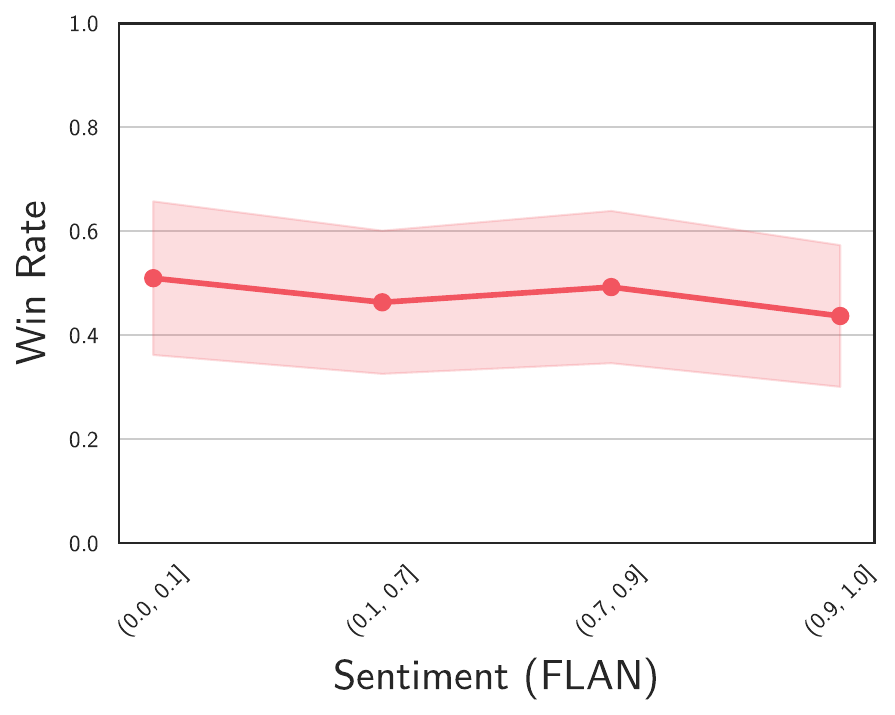}

}    
\subfigure[]{
    \includegraphics[width=0.315\textwidth]{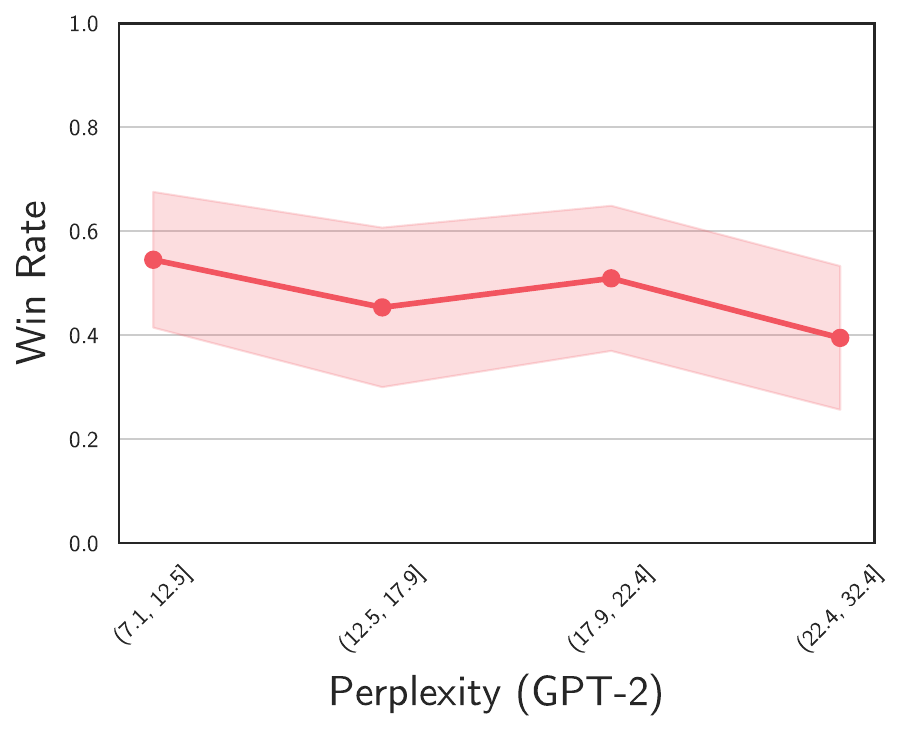}

}
\subfigure[]{
    \includegraphics[width=0.315\textwidth]{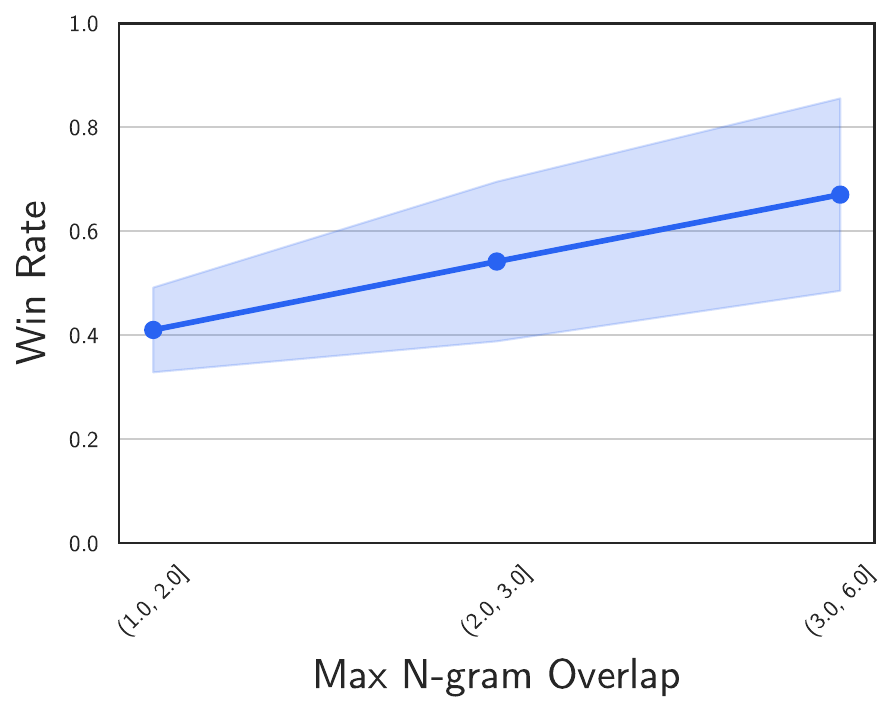}

}
\subfigure[]{
    \includegraphics[width=0.315\textwidth]{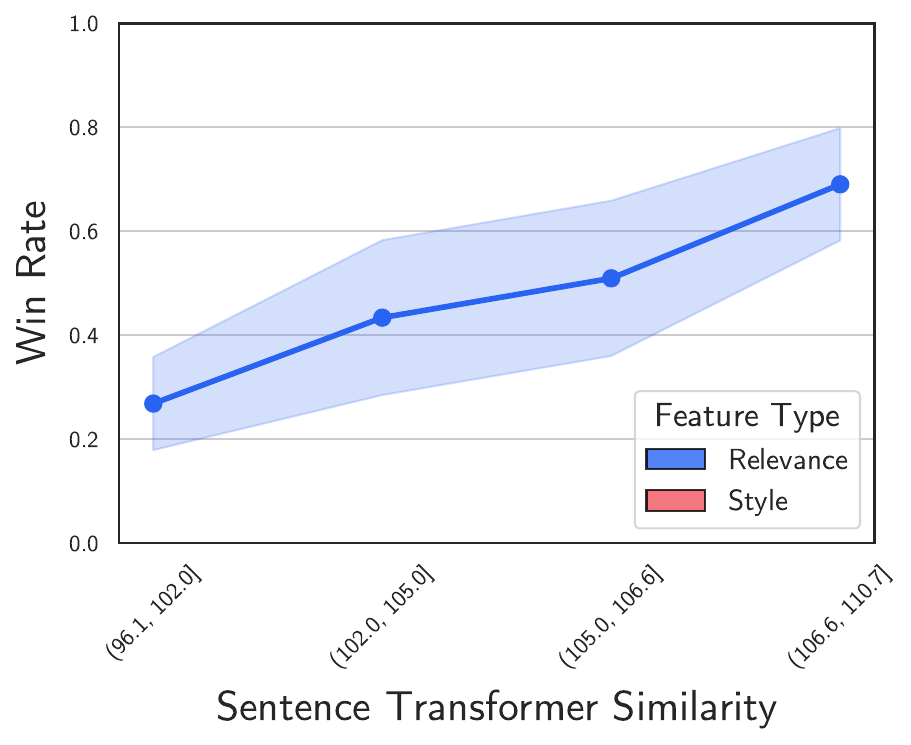}

}
\vspace{-0.3cm}
\caption{The analogous plots to Figure~\ref{fig:features_llamachat} except it is for \texttt{WizardLM 1.2 13B.} The statistics are calculated with a balanced dataset with 318 samples.}
\label{fig:features_wizardlm}
\end{figure*}

Figure~\ref{fig:features_claudeinstant}--\ref{fig:features_wizardlm} contain the analogous plots for Figure~\ref{fig:features_llamachat} across four other models.

\section{Expressing Convincingness in Isolation}
\label{sec:ungrounded_method}
We consider whether LLMs are able to express the convincingness of a paragraph in isolation. The model makes the rating using only the website. We prompt (Table~\ref{tab:ungrounded_prompt}) asking the model to rate the credibility of the website from a scale of one to five. The rating of the model is then determined by an average of the ratings, weighted by the probability of each label. Following~\citet{santurkar2023whose}, we calculate probabilities by exponentiating and normalizing the logits for "one" through "five". We also give the model examples of a "one" and "five" rating from C3~\cite{kakol2017c3}, a dataset for studying human credibility judgements. We use these few-shot examples as the model tends to be biased toward higher-ratings without them.

\section{Counterfactual Perturbations}
\label{app:perturb_descriptions}
\begin{figure*}[t]
\centering
\includegraphics[trim={0.1cm, 2.2cm, 0.1cm, 5.2cm}, clip, width=\textwidth]{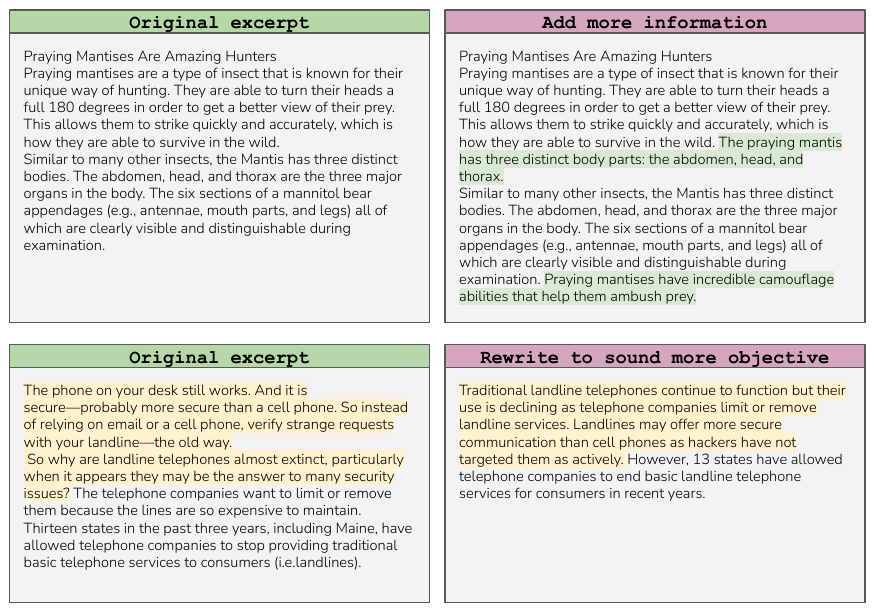}
\vspace{-0.70cm}
\caption{We use the \texttt{claude-v1-instant} model to make counterfactual edits to the text in order to change certain stylistic or relevancy features. Here, we show two examples of such perturbations.}
\label{fig:perturb_examples}
\end{figure*}

\begin{enumerate}[nosep]
    \item \texttt{Add Single Sentence}: We use \texttt{claude-v1-instant} to add a single sentence to make the stance of the text obvious. For example, for ``Does producing bottled water use more water than the bottle contains?'', we may add ``In fact, producing a single bottle of water uses more water than the bottle contains.''
    \item \texttt{Rewrite Relevance}: We alter the text with \texttt{claude-v1-instant} to make the text more relevant to the question.
    \item \texttt{Question Prefix}: We prefix the document with ``The following text is about the question: [question]''.
    \item \texttt{Keyword Stuffing}: We use \texttt{claude-v1-instant} to add additional sentences that use keywords related to the question.
    \item \texttt{Add More Info}: We use \texttt{claude-v1-instant} to add additional sentences of information that are unrelated to the question but related to the overall topic of the text. An example of this perturbation can be found in Figure~\ref{fig:perturb_examples}.
    \item \texttt{Add Science Reference}: We use \texttt{claude-v1-instant} to add scientific references to the text.
    \item \texttt{Add Contact Info}: We suffix the text with the name and phone number of a fake author.
    \item \texttt{Rewrite Confidence}: We use \texttt{claude-v1-instant} to make text sound more confident.
    \item \texttt{Rewrite Technical Language}: We use \texttt{claude-v1-instant} to make the text more technical.
    \item \texttt{Rewrite Objective}: We use \texttt{claude-v1-instant} to make the text more objective, e.g., Figure~\ref{fig:perturb_examples}.
\end{enumerate}

We include the prompts used to make these perturbations in Table~\ref{tab:perturbation_prompts}. We additionally measure the maximum n-gram overlap between the query and the perturbed samples to confirm that the altered paragraphs are, in fact, more relevant to the query. These results can be found in Table~\ref{tab:perturb_ngram}.

\begin{table*}[t]
\centering
\footnotesize
\begin{tabular}{lp{4.0in}}
\toprule
{\bf Perturbation} & {\bf Prompt}\\
\midrule
Rewrite Tech. Language & Please alter this text so that it uses technical language. Do not alter the underlying content of the text. Only alter the style of the text. Answer with nothing but the changed text. Only answer with the altered text.\\
\midrule
Add More Info & Please add 6 sentences to this text that are unrelated to "\{question\}", but related to the overall topic of the text. The added six sentences should not influence the answer of the "\{question\}" one way or the other. The added 6 sentences should be related to the overall topic of the text. The added 6 sentences should be sprinkled out throughout the text. Do not put ALL the sentences at the beginning or end. Answer with nothing but the changed text.\\
\midrule
Add Science Reference & Please alter this text so that it include references to scientific sources. Do not alter the underlying content of the text. Only alter the style of the text. Answer with nothing but the changed text. Only answer with the altered text.\\
\midrule
Rewrite Objective & Please alter this text so that it sounds more objective. Do not alter the underlying content of the text. Only alter the style of the text. Answer with nothing but the changed text. Only answer with the altered text.\\
\midrule
Rewrite Confidence & Please alter this text so that it is extremeley confident. Each sentence should be clear and unambiguous. Do not alter the underlying content of the text. Only alter the style of the text. Answer with nothing but the changed text. Only answer with the altered text.\\
\midrule
Keyword Stuffing & Please alter this text so that it uses keywords related to the question by additional sentences. Do not alter any existing sentences. Existing sentences should remain unaltered. Just add new sentences that use keywords related to the question. Do not alter the underlying content of the text. Only alter the style of the text. Answer with nothing but the changed text.\\
\midrule
Add Single Sentence & Please minimally alter this text so that the answer to the question "\{question\}" is obviously \{stance\}. You are only allowed to add a single sentence. Only add a single sentence and change nothing else. Answer with nothing but the changed text.\\
\midrule
Rewrite Relevance & Please alter this text so that it is more related to the question "\{question\}". Emphasize sentences that answer the question "\{question\}". Add a lot of keywords related to the question into the text. The text should use a lot of keywords related to the question "\{question\}". Answer with nothing but the changed text.\\
\bottomrule
\end{tabular}
\vspace{-0.1cm}
\caption{The prompts used to perform counterfactual perturbations on the text.}
\label{tab:perturbation_prompts}
\end{table*}

\begin{table*}[t]
\centering
\footnotesize
\begin{tabular}{llc}
\toprule
{\bf \shortstack{Perturbation}} & {\bf Type} & {\bf Overlap} \\
\midrule
Rewrite Relevance (Claude) & Relevance & 5.08 \\
Add Single Sentence (Claude) & Relevance & 3.64 \\
Question Prefix & Relevance & 6.89 \\
Keyword Stuffing (Claude) & Relevance & 2.47 \\
\midrule
Add Science Reference (Claude) & Style & 2.38 \\
Add More Info (Claude) & Style & 2.46 \\
Add Contact Info & Style & 2.41 \\
Rewrite Confidence (Claude) & Style & 2.17 \\
Rewrite Tech. Language (Claude) & Style & 1.97 \\
Rewrite Objective (Claude) & Style & 2.15 \\
\bottomrule
\end{tabular}

\caption{We calculate the maximum length n-gram that's in both the search query and the perturbed paragraph. We confirm that the paragraphs altered by Claude to be more relevant to the search query are also measured to be more relevant.}
\label{tab:perturb_ngram}
\end{table*}